\documentclass{article}

\usepackage[nonatbib,final]{neurips_2021_imagenet}

\usepackage[utf8]{inputenc} 
\usepackage[T1]{fontenc}    
\usepackage{hyperref}       
\usepackage{url}            
\usepackage{booktabs}       
\usepackage{amsfonts}       
\usepackage{nicefrac}       
\usepackage{microtype}      
\usepackage{rotating}
\usepackage{forest}

\usepackage{rotating}
\usepackage{graphicx}
\usepackage{amsmath}
\usepackage{amssymb}
\usepackage{amsthm}
\usepackage{dsfont}
\usepackage{multirow}

\usepackage{xcolor}
\usepackage{tikz}
\usetikzlibrary{shapes}
\usetikzlibrary{arrows, arrows.meta}
\usepackage{bm}
\usepackage{dirtytalk}
\usepackage{subcaption}

\def\@fnsymbol#1{\ensuremath{\ifcase#1\or \dagger\or \ddagger\or
   \mathsection\or \mathparagraph\or \|\or **\or \dagger\dagger
   \or \ddagger\ddagger \else\@ctrerr\fi}}
   
\title{Evaluating Adversarial Attacks on ImageNet:\\A Reality Check on Misclassification Classes}

\author{%
Utku Ozbulak$^\ast$\\ 
Ghent University, Belgium\\
\texttt{utku.ozbulak@ugent.be}
\And
Maura Pintor\thanks{Equal contribution.}\\
University of Cagliari, Italy \\
\texttt{maura.pintor@unica.it} \\
\AND
Arnout Van Messem\\
University of Li\`ege, Belgium\\
\texttt{Arnout.vanmessem@uliege.be} \\
\And
Wesley De Neve\\
Ghent University, Belgium\\
\texttt{wesley.deneve@ugent.be } \\
}

\begin{document}

\maketitle

\vspace{-0.5em}
\begin{abstract}
Although ImageNet was initially proposed as a dataset for performance benchmarking in the domain of computer vision, it also enabled a variety of other research efforts. Adversarial machine learning is one such research effort, employing deceptive inputs to fool models in making wrong predictions. To evaluate attacks and defenses in the field of adversarial machine learning, ImageNet remains one of the most frequently used datasets. However, a topic that is yet to be investigated is the nature of the classes into which adversarial examples are misclassified. In this paper, we perform a detailed analysis of these misclassification classes, leveraging the ImageNet class hierarchy and measuring the relative positions of the aforementioned type of classes in the unperturbed origins of the adversarial examples. We find that $71\%$ of the adversarial examples that achieve model-to-model adversarial transferability are misclassified into one of the top-5 classes predicted for the underlying source images. We also find that a large subset of untargeted misclassifications are, in fact, misclassifications into semantically similar classes. Based on these findings, we discuss the need to take into account the ImageNet class hierarchy when evaluating untargeted adversarial successes. Furthermore, we advocate for future research efforts to incorporate categorical information.
\end{abstract}

\vspace{-0.5em}
\section{Introduction}
\vspace{-1em}
Soon after its release, ImageNet~\cite{ILSVRC15:rus} became the de facto standard dataset for performance benchmarking in the field of computer vision, primarily thanks to the diverse set of images and classes it contains. This diversity allowed for research on various vision tasks, including, but not limited to, classification~\cite{Alexnet,VGG}, segmentation~\cite{SegNet,Fcn8s}, and localization~\cite{mask_rcnn,faster_rcnn}. Although the tasks put forward during the introduction of ImageNet were considered to be some of the hardest problems to address in the field of computer vision, a number of deep neural networks (DNNs) were, in recent years, able to achieve super-human results on many of these challenges, thus effectively \say{solving} the aforementioned problems~\cite{dosovitskiy2021an}. However, research efforts that make use of ImageNet are not limited to the performance-oriented tasks mentioned before. Indeed, thanks to the diverse set of images it contains, ImageNet enabled a large number of research efforts beyond its initial scope, allowing researchers to experiment with model interpretability~\cite{grad_cam,guided_backprop}, model calibration~\cite{guo2017calibration}, object relations~\cite{RussakovskyFeiFei}, fairness~\cite{yang2019fairer}, and many other topics.

One research field that was enriched by the availability of ImageNet is the field of study that focuses on adversarial examples. In this context, the term \say{adversarial examples} refers to meticulously created data points that come with a malicious intent, aimed at deceiving models that are performing a pre-defined task, steering the prediction outcome in favor of the adversary~\cite{biggio2013evasion,LBFGS}. Although adversarial examples are a threat for predictive models in domains other than the domain of computer vision~\cite{carlini2018audio,ozbulak2021investigating}, the latter is acknowledged to be the one that suffers the most from adversarial examples, since an adversarial example created from a genuine image, through the use of adversarial perturbation, often looks the same as its unperturbed counterpart~\cite{Goodfellow-expharnessing,mcdaniel2016machine}. This makes it, in most cases, impossible to detect adversarial examples by visually inspecting images.

Although the vulnerability of DNNs to adversarial examples in the image domain was originally mostly evaluated through the usage of two datasets, namely MNIST~\cite{lecun1998gradient} and CIFAR~\cite{CIFAR}, the authors of~\cite{DBLP:journalsCarliniW17} revealed that methods derived through the usage of one of these datasets do not necessarily generalize to other datasets. In particular, compared to ImageNet, both of the aforementioned datasets contain images with a smaller resolution and a lower number of classes. As a result, most of the research efforts in recent years started to favor ImageNet over MNIST and CIFAR~\cite{croce2019sparse,guo2019simple_black_black_box,su2018adversarial_activation_functions,xu2018structured_new_local_adv_attack}.

From the perspective of adversarial evaluation, ImageNet does not only allow for most, if not all, of the research work that was performed using the previously mentioned datasets, it also enables a wide range of additional research topics in the area of adversariality, such as investigations with regards to regional perturbation~\cite{LAVAN}, color channels~\cite{shamsabadi2020colorfool,xu2018structured_new_local_adv_attack}, and defenses that use certain properties of natural images~\cite{total_variation_defense}. However, as demonstrated in this paper, ImageNet has a major shortcoming when it comes to evaluating adversarial attacks, especially in model-to-model transferability scenarios: a large number of synsets/classes in ImageNet are semantically highly similar to one another.

Different from previous research efforts that mostly focus on generating more effective adversarial perturbations or evaluating adversarial defenses, we investigate a topic that is yet to be touched upon: untargeted misclassification classes for adversarial examples. Specifically, with the help of two of the most frequently used adversarial attacks and seven unique DNN architectures, including two recently proposed vision transformer architectures, we present a large-scale study that solely focuses on model-to-model adversarial transferability and misclassification classes in the context of ImageNet, resulting in the following contributions:

\quad\textbullet\,\,In model-to-model transferability scenarios, we demonstrate that a large portion of adversarial examples are classified into the top-5 predictions obtained for their source image counterparts.

\quad\textbullet\,\,With the help of the ImageNet class hierarchy, we show that adversarial examples created from certain synset collections are mostly misclassified into classes belonging to the same collections (e.g., a dog breed is misclassified as another dog breed).

\quad\textbullet\,\,Interestingly, we can make the two aforementioned observations consistently for all of the evaluated models, as well as for both adversarial attacks. As a result, we discuss the necessity of evaluating misclassification classes when experimenting with adversarial attacks and untargeted misclassification in the context of ImageNet.


\vspace{-0.5em}
\section{Adversarial attacks}
\vspace{-1em}
Given an $M$-class classification problem, a data point $\bm{x}\in \mathbb{R}^k$ and its categorical association $\bm{y} \in \mathbb{R}^M$ associated with a correct class $k$ ($y_k = 1$ and $y_m = 0 \,, \forall \, m \in \{0,\ldots, M\} \char`\\ \{k\}$) are used to train a machine learning model represented by $\theta$. Let $g(\theta, \bm{x}) \in \mathbb{R}^M$ represent the prediction (logit) produced by the model $\theta$ and a data point $\bm{x}$. This data point is then assigned to the class that contains the largest output value $G(\theta, \bm{x}) = \arg \max (g(\theta, \bm{x}))$. When $G(\theta, \bm{x}) = \arg \max (\bm{y})$, this prediction is recognized as the correct one. For the given setting, a perturbation $\Delta$ bounded by an $L_p$ ball centered at $\bm{x}$ with radius $\epsilon$ is said to be an \textit{adversarial perturbation} if $G(\theta, \bm{x}) \neq G(\theta, \bm{x} + \Delta)$. In this case, $\hat{\bm{x}} = \bm{x} + \Delta$ is said to be an \textit{adversarial example}.

Adversarial examples can be highly \textit{transferable}: an adversarial sample that fools a certain classifier can also fool completely different classifiers that have been trained for the same task~\cite{cheng2019improvinge_black_black_box,demontis2019transferability,DBLP:journals/corr/PapernotMG16}. This property, which is called transferability of adversarial examples, is a popular metric for assessing the effectiveness of a particular attack. Let $\theta_1$ and $\theta_2$ represent two DNNs and let $\bm{x}$, $k$, and $\hat{\bm{x}}_1$ be a genuine image, the correct class of this image, and a corresponding adversarial example, respectively, with the adversarial example generated from this genuine image using an attack that targets a class $c$ by leveraging the DNN represented by $\theta_1$. If $G(\theta_1, \hat{\bm{x}}_1) = G(\theta_2, \hat{\bm{x}}_1) = c$ and $G(\theta_{\{1,2\}}, \bm{x}) = k$, then the adversarial example is said to have achieved \textit{targeted adversarial transferability} to the model $\theta_2$. If $G(\theta_1, \hat{\bm{x}}_1) =c$ but $G(\theta_2, \hat{\bm{x}}_1) \notin \{c, k\}$, the adversarial example in question is classified into a class that is different than the targeted one ($c$) and the correct one ($k$). In cases like this, an adversarial example is said to have achieved \textit{untargeted adversarial} transferability.

In the context of ImageNet, the success of targeted transferability for adversarial examples is known to be abysmally lower compared to the success of untargeted transferability~\cite{su2018robustness_18_imagenet_models_evaluation}. As a result, many studies that propose a novel attack or perform a large-scale analysis of model-to-model transferability use untargeted transferability when showcasing the effectiveness of attacks, without evaluating the classes that adversarial examples are classified into~\cite{croce2019sparse,guo2019simple_black_black_box,xu2018structured_new_local_adv_attack}. Therefore, in this work, we investigate the success of untargeted adversarial transferability and the characteristics of misclassification classes.

\begin{figure}[t!]
    \centering
    \includegraphics[width=0.4\linewidth]{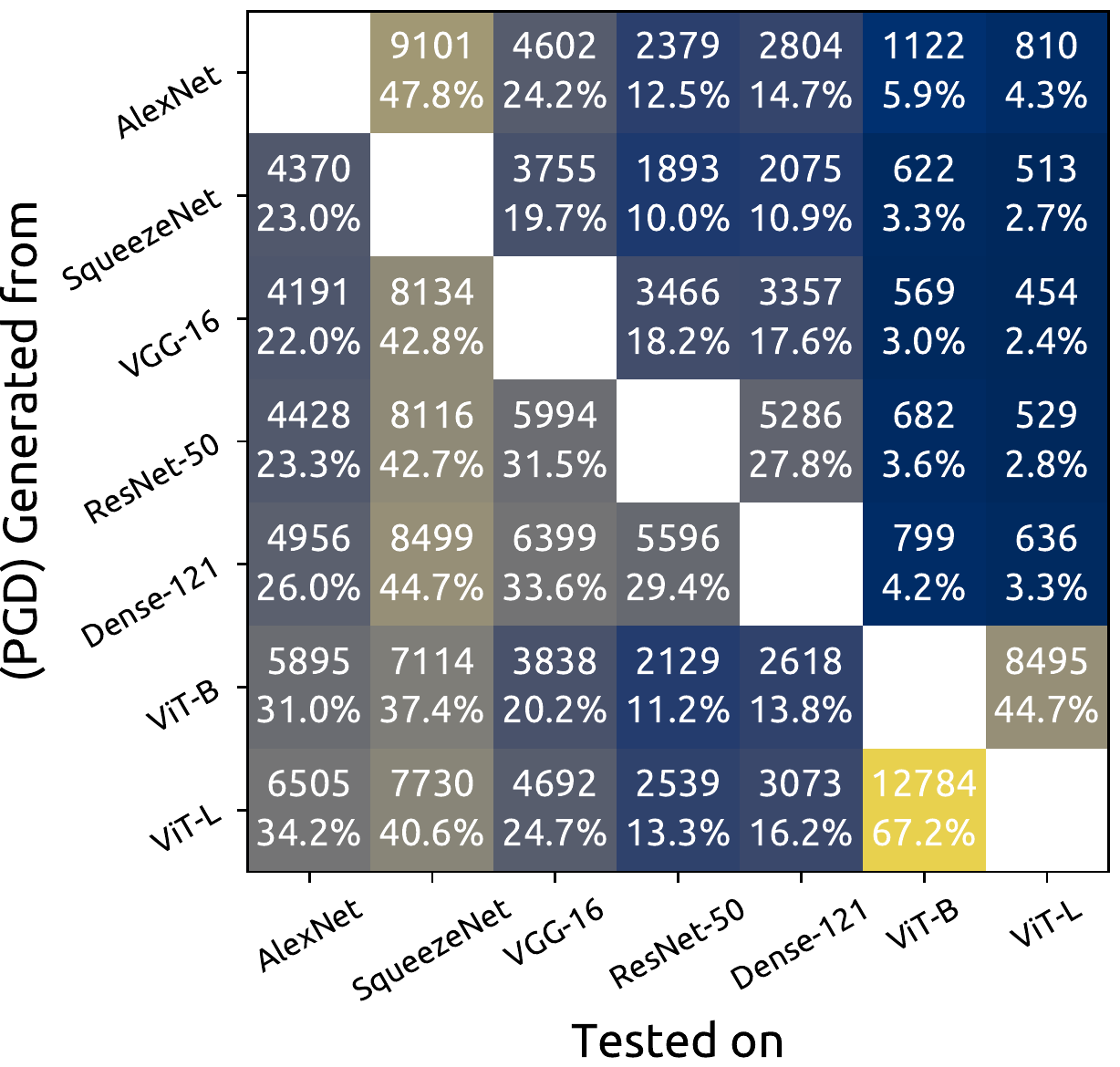}\quad\quad\quad
    \includegraphics[width=0.4\linewidth]{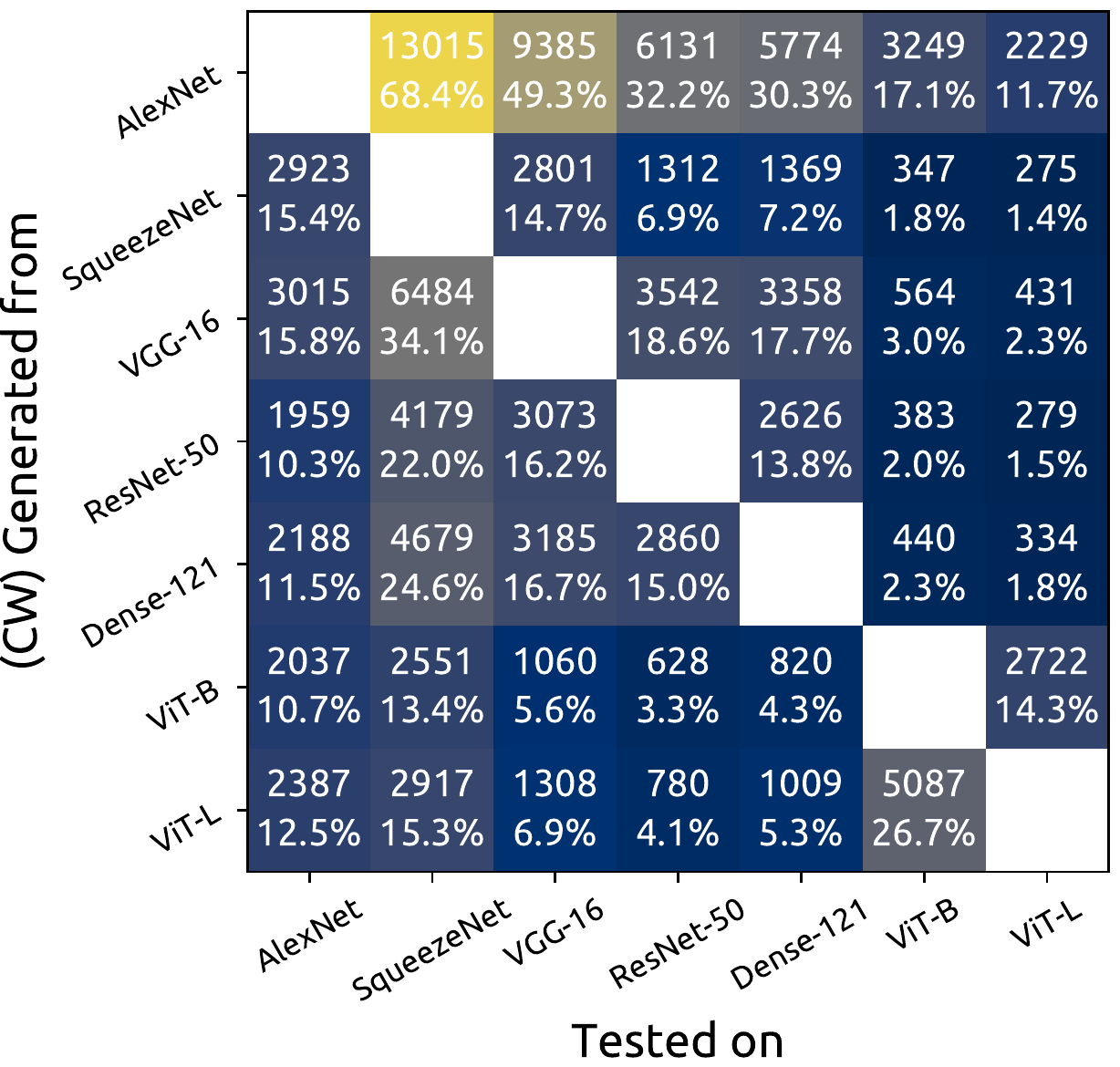}
    \caption{Number (percentage) of source images that became adversarial examples with PGD (\textit{left}) and CW (\textit{right}). Adversarial examples are generated by the models listed along the $y$-axis and tested by the models listed along the $x$-axis.}
    \label{fig:transferability_matrix_untargeted}
\end{figure}

\vspace{-1em}
\section{Methodology}
\vspace{-0.5em}
\textbf{Models}\,\textendash\,In order to evaluate a variety of model-to-model adversarial transferability scenarios, we employ the following architectures: AlexNet~\cite{Alexnet}, SqueezeNet~\cite{squeezenet}, VGG-16~\cite{VGG}, ResNet-50~\cite{resnet}, and DenseNet-121~\cite{densenet}, as well as two recently proposed vision transformer architectures, namely ViT-Base$/16-224$ and ViT-Large$/16-224$~\cite{dosovitskiy2021an}.

\textbf{Data}\,\textendash\,For our adversarial attacks (see further in this section), we use images from the ImageNet validation set as inputs. Hereafter, these unperturbed input images will be referred to as \textit{source images}. In order to perform a trustworthy analysis of adversarial transferability, we ensure that all source images are correctly classified by all employed models. To that end, we filter out all images incorrectly classified by at least one model, leaving us with $19,025$ source images to work with.

\textbf{ImageNet hierarchy}\,\textendash\,Classes in ImageNet are organized according to the WordNet hierarchy~\cite{WordNet,ILSVRC15:rus}, grouping classes into various collections depending on their semantic meaning. We use the aforementioned hierarchy in order to measure intra-collection adversarial misclassifications. In that respect, an intra-collection misclassification is when an adversarial example created from a source image that belongs to a class under a collection is misclassified into a class under the same collection (e.g., an image belonging to a cat breed misclassified as another breed of cat is an intra-collection misclassification for the \textit{Feline} collection). More details about the ImageNet hierarchy are given in the supplementary material (see Figure I).

\textbf{Attacks}\,\textendash\,We use the adversarial examples generated for our previous study~\cite{ozbulak_selection}, where those adversarial examples are generated using two of the most commonly used attacks: Projected Gradient Descent (PGD)~\cite{PGD_attack} and Carlini \& Wagner's attack (CW)~\cite{CW_Attack}. 

PGD can be seen as a generalization of $L_{\infty}$ attacks~\cite{Goodfellow-expharnessing,IFGS}, aiming at finding an adversarial example $\hat{\bm{x}}$ that satisfies $||\hat{\bm{x}} - \bm{x}||_{\infty} < \epsilon$. The adversarial example is iteratively generated as follows:
\begin{align}
\hat{\bm{x}}^{(n+1)} =  \Pi_{\epsilon}\Big(\hat{\bm{x}}^{(n)} - \alpha \ \text{sign} \big(\nabla_x J(g(\theta, \hat{\bm{x}}^{(n)})_c)  \big)\Big) \,,
\end{align}
with $\hat{\bm{x}}^{(1)} = \bm{x}$, $c$ the selected class, and $J(\cdot)$ the cross-entropy loss. We use PGD with $50$ iterations and set $\epsilon$ to $38/255$. We adopt this constraint as the maximum perturbation-size bound in order to be able to produce a large number of adversarial examples that achieve model-to-model transferability.

CW, on the other hand, is a complex attack that incorporates $L_2$ norm minimization:
\begin{align}
\text{miminize} \quad & ||\bm{x} - (\bm{x} + \Delta)||_{2}^{2} +  \; f(\bm{x} + \Delta)\,.
\end{align}

In the paper introducing CW~\cite{CW_Attack}, multiple loss functions (i.e., $f$) are discussed. However, in later works, the creators of CW prefer to make use of the loss function that is constructed as follows:
\begin{align} \label{eq:CW_2}
f (\bm{x}) = \max \big(  \max \{ g(\theta, \bm{x})_i : i\neq c\} - g(\theta, \bm{x})_c, -  \kappa \big) \,,
\end{align}
where this loss compares the predicted logit value of target class $c$ with the predicted logit value of the next-most-likely class $i$. The constant $\kappa$ can be used to adjust the \textit{strength} of the produced adversarial examples (for our experiments, we use $\kappa=20$ and the settings described in~\cite{CW_Attack} and~\cite{ozbulak_selection}).

We keep executing the attacks until a source image becomes an adversarial example or until the attacks reach a maximum number of iterations. At each iteration, we examine whether or not the images under consideration became adversarial examples for the aforementioned models.

\begin{figure}[t!]
\begin{tikzpicture}
\centering
\node[inner sep=0pt] (a) at (0, 0)
{\includegraphics[width=0.89\linewidth]{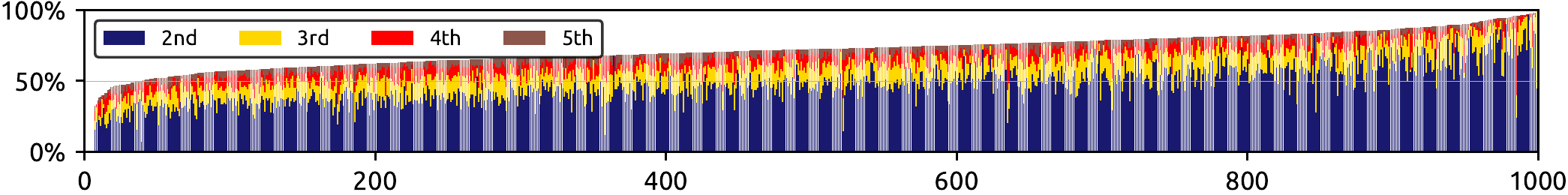}};
\node[align=center,rotate=90] at (-6.75, 0) {\scriptsize Adversarial examples\\\scriptsize  misclassified into\\\scriptsize top-K classes};

\node[align=left] at (0.25, -1) {\scriptsize ImageNet class (sorted according to y-axis)};
\end{tikzpicture}
\caption{Number of adversarial examples, given per class, that are classified into the top$-\{2,3,4,5\}$ classes predicted for their underlying source images.}
\label{fig:top5_misclassifications}
\end{figure}

\section{Experiments}
\vspace{-0.5em}
Leveraging the attacks described above and through the usage of $19,025$ source images that are correctly classified by the models employed, we create $289,244$ adversarial examples, where $173,549$ of those adversarial examples are generated with PGD and $115,695$ with CW. Detailed untargeted model-to-model transferability successes of those adversarial examples can be found in Figure~\ref{fig:transferability_matrix_untargeted}. 

To investigate misclassifications made into semantically similar classes, we first have a look at the adversarial examples that are misclassified into classes that lie in the top-5 positions of their source image predictions, where the four remaining classes, apart from the first one, are the classes that were deemed to be the most-likely prediction classes by the model under consideration, with the first one being the correct classification. Doing so, we provide Figure~\ref{fig:top5_misclassifications}, with this figure displaying, for each class, the percentage of adversarial examples that had their predictions changed into one of the top-5 classes as described above. Specifically, we observe that $215,717$ (approximately $71\%$) adversarial examples are predicted into one of the top-5 predictions of their unperturbed source images, where these classes in the top-5 are often highly similar to the correct predictions for the source images the adversarial examples are generated from (see Figure~II in the supplementary material).

Although this graph hints that a large portion of untargeted adversarial transferability successes are (plausible) misclassifications rather than adversarial successes, on its own, it does not provide enough evidence to make such a claim. In order to solidify this observation, we expand on misclassifications and utilize the ImageNet class hierarchy. In Table~\ref{tbl:main_pgd_cw_short}, we provide the count and the percentage of adversarial examples that are originating from a number of collections and their intra-collection misclassification rates for a number of collections under the \textit{Organism} branch of the hierarchy. Table~\ref{tbl:main_pgd_cw_short} represents the aforementioned measurements for all adversarial examples that achieved adversarial transferability to any of the models and with any attack.

Naturally, the larger the collection, the higher the intra-collection misclassification rate will be. For example, a source image taken from the \textit{Organism} collection has $409$ other classes that may contribute to intra-collection misclassification. However, even for smaller, more granular collections such as the \textit{Bird} collection, which only contains $59$ classes, we observe that adversarial examples are more-often-than-not misclassified into the classes in the same collection. Furthermore, a number of collections such as \textit{Canine}, \textit{Bird}, \textit{Reptilian}, and \textit{Arthropod} stand out among other collections for having remarkably high intra-collection misclassification rates. For example, $84\%$ of all adversarial examples that originate from a canine (i.e., dog) image are misclassified as another breed of canine.


In Table~\ref{tbl:main_pgd_cw_short}, we also provide misclassifications into the top-3 and the top-5 classes for adversarial examples that are originating from source images taken from individual collections. As can be seen, the observations we made when evaluating all adversarial examples also hold true for individual collections, where most of the adversarial examples in those collections have a misclassification rate of about $60\%$ and $70\%$ for the top-3 and the top-5 classes, respectively. To make matters worse, we can even see trends similar to the aforementioned observations when we filter adversarial examples for individual attacks and when we investigate misclassifications on a model-to-model basis, demonstrating that our observations are not specific to a single model or to one of the attacks. Extended results covering more collections and individual models/attacks can be found in the supplementary material (Table~I to Table~V).


\begin{table}[t!] 
\centering
\caption{For the adversarial examples that achieved model-to-model transferability, intra-collection misclassifications and misclassifications into the top-\{3,5\} prediction classes in the target models are provided. The results for the adversarial examples are grouped into collections according to the classes of their source image origins.}
\scriptsize
\begin{tabular}{llccccccc}
\cmidrule[0.25pt]{1-9}
\multirow{4}{*}{\shortstack{Hierarchy}} & 
\multirow{4}{*}{\shortstack{Collection}} & 
\multirow{4}{*}{\shortstack{Classes\\in collection}} & 
\multirow{4}{*}{\shortstack{Source\\images\\in collection}} & 
\multirow{4}{*}{\shortstack{Adversarial\\examples\\originating\\from collection}} & 
\multicolumn{2}{c}{\multirow{3}{*}{\shortstack{Intra-collection\\misclassifications}}} &
\multicolumn{2}{c}{\multirow{3}{*}{\shortstack{Misclassification\\into top-K\\classes}}} \\
~ & ~ & ~ \\
~ & ~ & ~ \\
\cmidrule[0.25pt]{6-9}
~ & ~ & ~ & ~ & ~ &  Count & \%  & Top-3 & Top-5 \\
\cmidrule[0.25pt]{1-9}
~ & All & 1000 & 19,025 & 289,244 & 289,244 & 100.0\% & 59.6\% & 71.1\%  \\
\cmidrule[0.25pt]{1-9}
1 & Organism & 410 & 9,390 & 147,621 & 132,865 & \bf 90.0\% & 61.2\% & 72.8\%  \\
1.1 & Creature & 398 & 9,009 & 143,996 & 130,409 & \bf 90.6\% & 61.4\% & 73.1\%  \\
1.1.1 & Domesticated animal & 123 & 2,316 & 50,036 & 41,978 & \bf 83.9\% & 63.4\% & 75.6\%  \\
1.1.2 & Vertebrate & 337 & 7,692 & 126,913 & 112,828 & \bf 88.9\% & 61.3\% & 73.2\%  \\
1.1.2.1 & Mammalian & 218 & 4,665 & 89,004 & 76,351 & \bf 85.8\% & 61.4\% & 73.5\%  \\
1.1.2.1.1 & Primate & 20 & 475 & 9,333 & 5,301 & \bf 56.8\% & 58.9\% & 70.4\%  \\
1.1.2.1.2 & Hoofed mammal & 17 & 419 & 6,206 & 2,751 & 44.3\% & 58.4\% & 71.6\%  \\
1.1.2.1.3 & Feline & 13 & 319 & 3,895 & 1,998 & \bf 51.3\% & 64.3\% & 75.9\%  \\
1.1.2.1.4 & Canine & 130 & 2,502 & 53,294 & 45,089 & \bf 84.6\% & 63.5\% & 75.7\%  \\
1.1.2.2 & Aquatic vertebrate & 16 & 366 & 5,355 & 2,383 & 44.5\% & 65.0\% & 75.6\%  \\
1.1.2.3 & Bird & 59 & 1,937 & 22,402 & 15,993 & \bf 71.4\% & 59.8\% & 71.3\%  \\
1.1.2.4 & Reptilian & 36 & 547 & 7,635 & 4,795 & \bf 62.8\% & 63.8\% & 75.2\%  \\
1.1.2.4.1 & Saurian & 11 & 188 & 2,416 & 1,050 & 43.5\% & 58.4\% & 71.1\%  \\
1.1.2.4.2 & Serpent & 17 & 223 & 3,202 & 1,700 & \bf 53.1\% & 67.0\% & 77.1\%  \\
1.1.3 & Invertebrate & 61 & 1,317 & 17,083 & 10,698 & \bf 62.6\% & 61.9\% & 72.3\%  \\
1.1.3.1 & Arthropod & 47 & 1,018 & 13,200 & 8,863 & \bf 67.1\% & 63.1\% & 73.5\%  \\
1.1.3.1.1 & Insect & 27 & 652 & 7,850 & 4,468 & \bf 56.9\% & 59.9\% & 70.5\%  \\
1.1.3.1.2 & Arachnoid & 9 & 189 & 2,824 & 1,476 & \bf 52.3\% & 69.7\% & 79.5\%  \\
1.1.3.1.3 & Crustacean & 9 & 137 & 2,035 & 955 & 46.9\% & 70.0\% & 80.1\%  \\
\cmidrule[1pt]{1-9}
\end{tabular}
\label{tbl:main_pgd_cw_short}
\vspace{-2em}
\end{table}

\vspace{-0.5em}
\section{Conclusions and outlook}
\vspace{-0.5em}
In the context of a classification problem, what differentiates an adversarial success from a plausible misclassification? If an adversarial example is misclassified into a class that is highly similar to the class of its unperturbed origin, should it still be considered an adversarial success? In this case, how should we measure the similarity between the classes? The aforementioned questions are not trivial to answer, and different answers may find different logical explanations depending on the context of the evaluation performed. However, given that the threat of adversarial examples is evaluated from the perspective of security, does a semantically similar misclassification that has been made in the context of ImageNet (e.g., a brown dog breed misclassified as another brown dog breed) carry the same weight as a lethal misclassification in the context of self-driving cars (e.g., a road sign misclassification leading to an accident)?

Finding answers to the questions presented above requires meticulous investigations on the topic of misclassification classes, where these investigations should involve various threat scenarios, similar to the work presented in~\cite{schwinn2021exploring,zhang2020understanding,zhao2020success}. In this paper, we took one of the first steps in analyzing misclassification classes in the context of ImageNet, with the help of large-scale experiments and the ImageNet class hierarchy, showing that a large number of untargeted adversarial misclassifications in model-to-model transferability scenarios are, in fact, plausible misclassifications. In particular, we observe that categories under the \textit{Organism} branch have considerably high intra-collection misclassifications compared to classes in the \textit{Artifact} branch. To aid future work on this topic in the context of ImageNet, we share an easy-to-use class hierarchy of ImageNet, as well as other resources, in the following repository: {\color{magenta}\url{https://github.com/utkuozbulak/imagenet-adversarial-image-evaluation}}.

\clearpage

\bibliographystyle{abbrv}
\bibliography{main}

\appendix

\clearpage
\newpage

\begin{center}
\begin{LARGE}
Supplementary Materials for:\\Evaluating Adversarial Attacks on ImageNet:\\A Reality Check on Misclassification Classes
\end{LARGE}
\end{center}

\begin{figure}[h!]
\centering
\begin{subfigure}[t]{1\textwidth}
\centering
\scriptsize
\begin{forest}
  for tree={l+=0.15cm} 
  [\normalsize\textbf{All}
    [Artifact[522]]
    [Organism[410]]
    [Natural object[17]]
    [Vegetable[13]]
    [Nutrition[10]]
    [Geological formation[10]]
    [Fungus[7]]
    [Beverage[4]]
  ]
\end{forest}
\caption{\normalsize Main branches of the ImageNet class hierarchy and the number of classes within those branches.}
\end{subfigure}
\\
\vspace{2em}
\begin{subfigure}[t]{1.1\textwidth}
\centering
\scriptsize
\begin{forest}
  for tree={l+=0.15cm} 
    [\normalsize\textbf{Organism}
        [Creature
            [Invertebrate[Arthropod[Insect][Arachnoid][Crustacean]]]
            [Domestic animal]
            [Vertebrate
                [Mammalian[Primate][Canine][Feline][Hoofed animal]]
                [Bird][Aquatic vertebrate][Reptilian[Saurian][Serpent]]
            ]
        ]
    ]
\end{forest}
\caption{\normalsize  ImageNet \textit{Organism} sub-tree.}
\end{subfigure}
\\
\vspace{2em}
\begin{subfigure}[t]{1\textwidth}
\centering
\scriptsize
\begin{forest}
  for tree={l+=0.15cm} 
    [\normalsize{\textbf{Artifact}}
        [Covering[Protective covering]]
        [Instrumentation]
        [Structure[Building]]
        [Commodity[Consumer good[Durable][Clothing[Garment]]]]
    ]
\end{forest}
\caption{\normalsize  ImageNet \textit{Artifact} sub-tree.}
\end{subfigure}
\\
\vspace{2em}
\begin{subfigure}[t]{1\textwidth}
\centering
\scriptsize
\begin{forest}
  for tree={l+=0.15cm}
    [\normalsize{\textbf{Instrumentation}}
        [Container[Vessel][Wheeled vehicle[Self-propelled vehicle[Motor vehicle]]]]
        [Transport[Vehicle[Air craft][Water craft]]]
        [Furnishing]
        [Device
            [Instrument[Weapon][Measuring instrument]]
            [Musical  instrument[Stringed][Wind]]
            [Machine]
            [Mechanism]
            [Equipment[Electronic][Game]]
        ]
        [Implement]
    ]
\end{forest}
\caption{\normalsize  ImageNet \textit{Instrumentation} sub-tree under \textit{Artifact} branch.}
\end{subfigure}
\caption{The ImageNet class hierarchy: (a) main branches and the number of classes that lie in those branches, (b) view of \textit{Organism} sub-tree, (c) view of \textit{Artifact} sub-tree, and (d) view of \textit{Instrumentation} sub-tree.}
\label{fig:ImageNetHier_mai}
\end{figure}
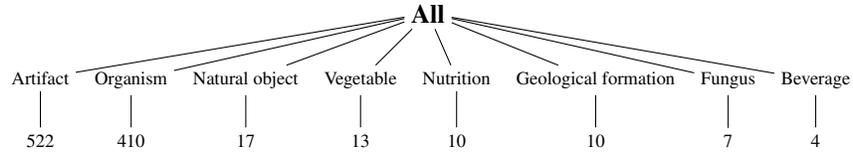
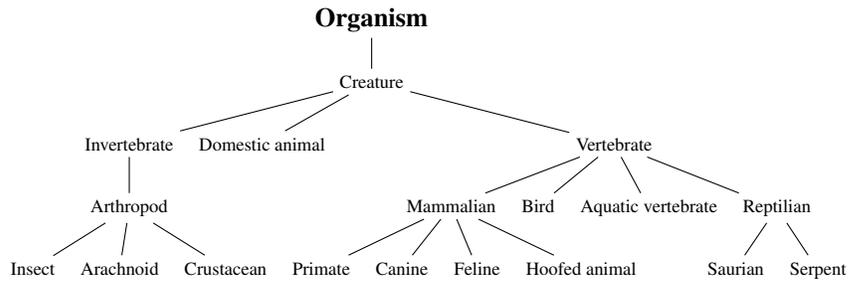
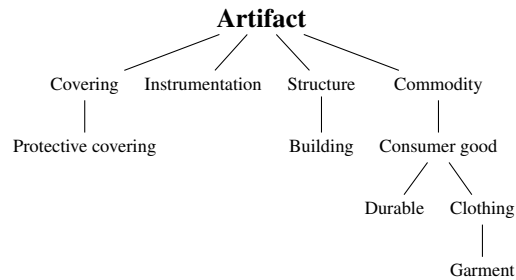
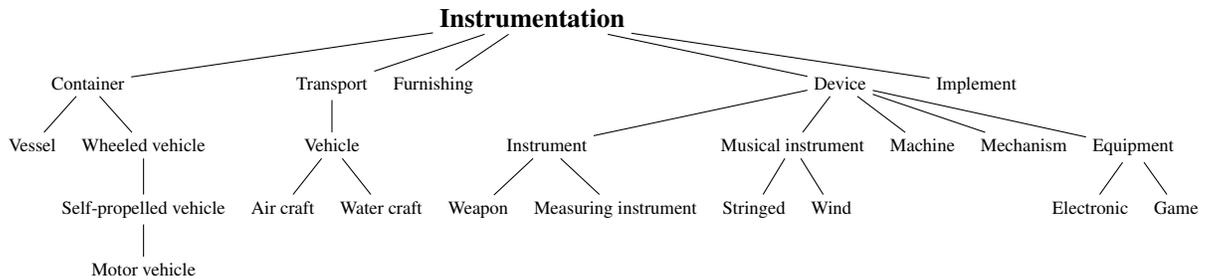

\clearpage
\begin{figure}[t!]
\centering
\includegraphics[width=0.49\linewidth]{transferability_matrix/PGD_Detailed_all_ims_transferability_matrix.pdf}\,
\includegraphics[width=0.49\linewidth]{transferability_matrix/CW_Detailed_all_ims_transferability_matrix.pdf}
\\
\vspace{3em}
\includegraphics[width=0.49\linewidth]{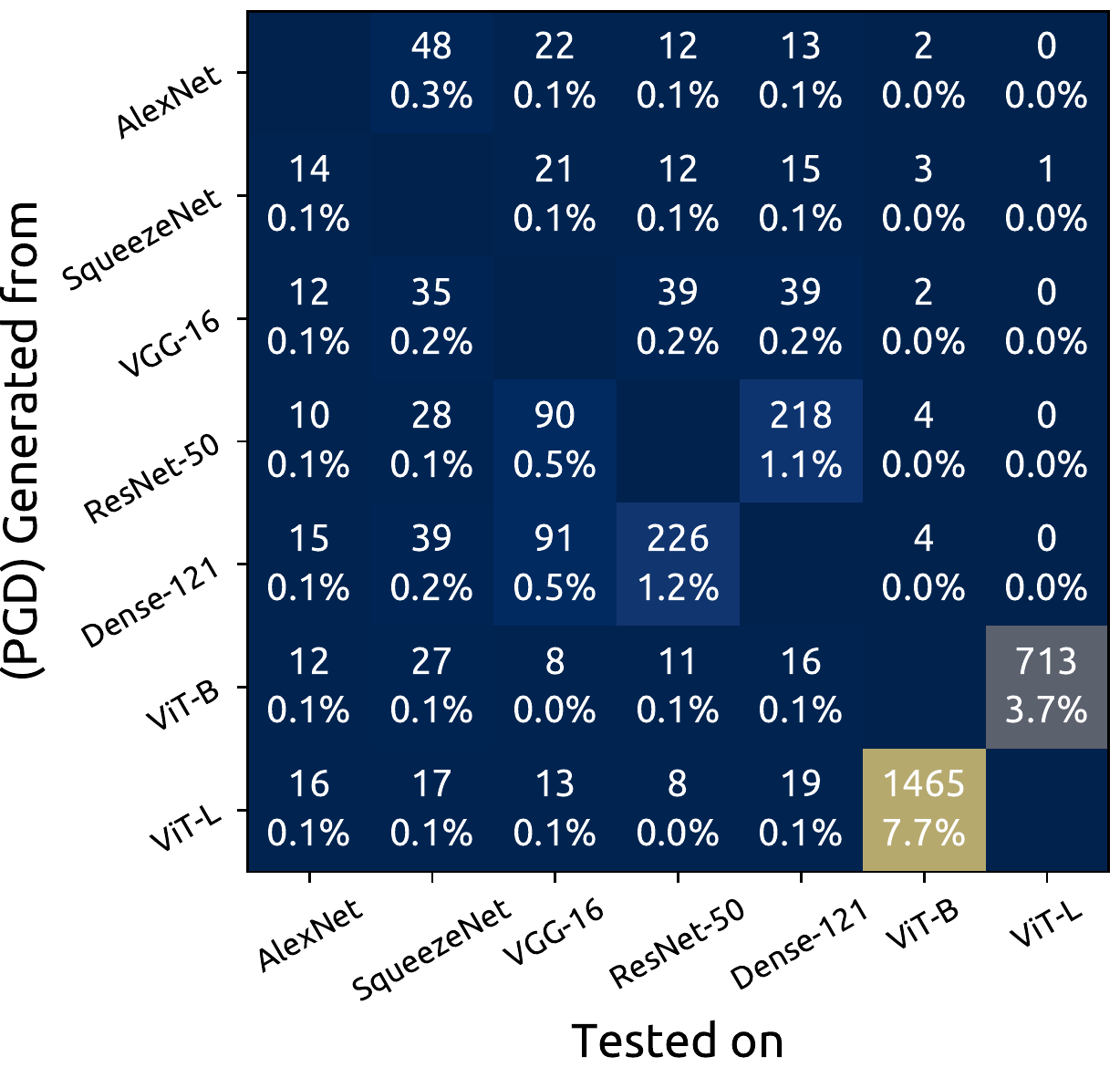}\,
\includegraphics[width=0.49\linewidth]{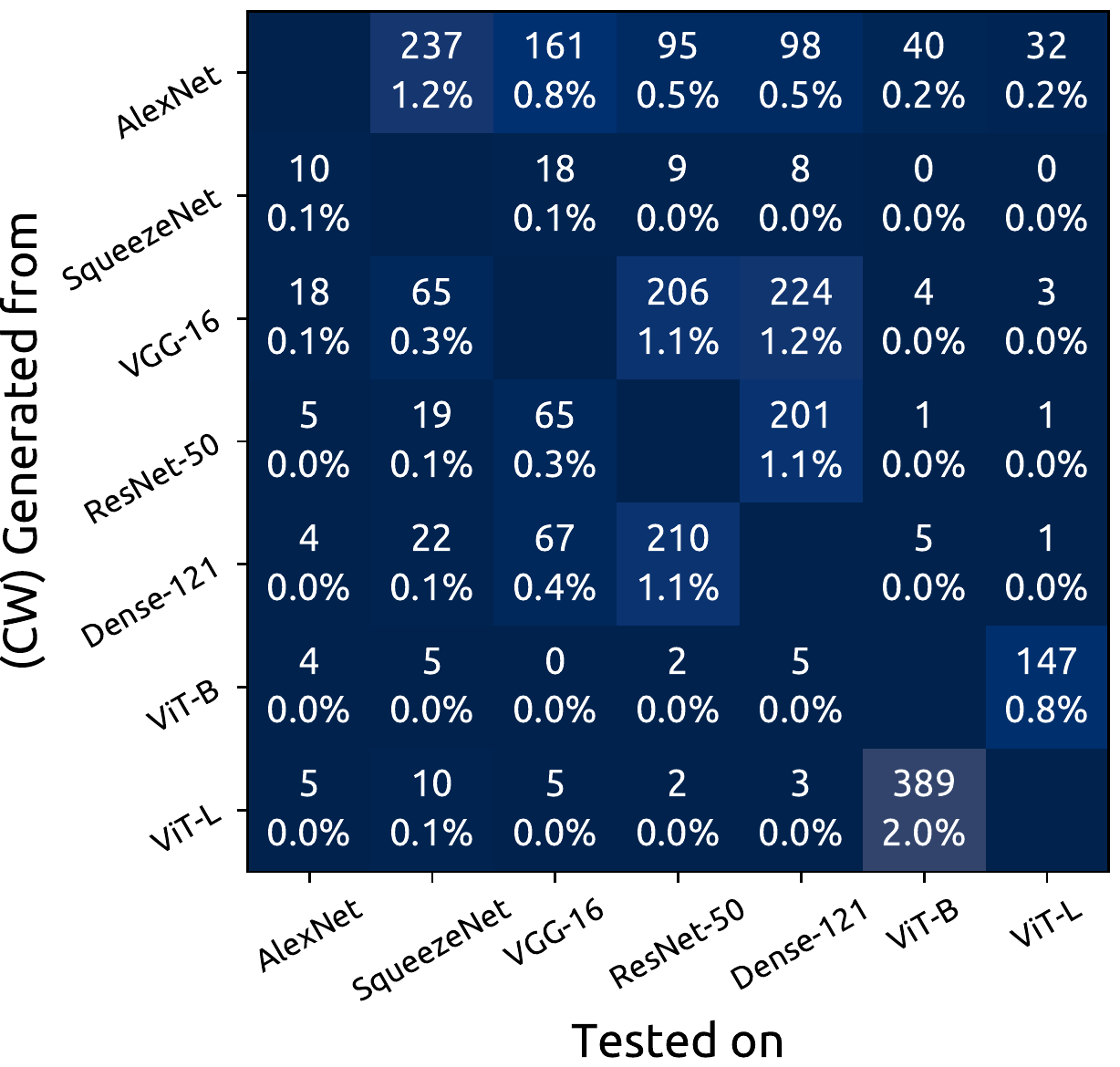}
\vspace{2em}
\caption{Number (percentage) of source images that became adversarial examples with PGD (\textit{left}) and CW (\textit{right}). Adversarial examples are generated by the models listed along the $y$-axis and tested by the models listed along the $x$-axis. The two figures at the top display untargeted transferability successes, whereas the two figures at the bottom display targeted transferability successes.}
\label{fig:transferability_matrix_detailed}
\end{figure}

\clearpage
\begin{table}[htbp!] 
\centering
\caption{For the adversarial examples that achieved model-to-model transferability and that have been created with 
\textbf{PGD} and \textbf{CW}, intra-collection misclassifications and misclassifications into the top-\{3,5\} prediction classes in the target models are provided. The results for the adversarial examples are grouped into collections according to the classes of their source image origins.}
\scriptsize
\begin{tabular}{llccccccc}
\cmidrule[0.25pt]{1-9}
\multirow{4}{*}{\shortstack{Hierarchy}} & 
\multirow{4}{*}{\shortstack{Collection}} & 
\multirow{4}{*}{\shortstack{Classes\\in collection}} & 
\multirow{4}{*}{\shortstack{Source\\images\\in collection}} & 
\multirow{4}{*}{\shortstack{Adversarial\\examples\\originating\\from collection}} & 
\multicolumn{2}{c}{\multirow{3}{*}{\shortstack{Intra-collection\\misclassifications}}} &
\multicolumn{2}{c}{\multirow{3}{*}{\shortstack{Misclassification\\into top-K\\classes}}} \\
~ & ~ & ~ \\
~ & ~ & ~ \\
\cmidrule[0.25pt]{6-9}
~ & ~ & ~ & ~ & ~ &  Count & \%  & Top-3 & Top-5 \\
\cmidrule[0.25pt]{1-9}
~ & All & 1000 & 19,025 & 289,244 & 289,244 & 100.0\% & 59.6\% & 71.1\%  \\
\cmidrule[0.25pt]{1-9}
1 & Organism & 410 & 9,390 & 147,621 & 132,865 & \bf 90.0\% & 61.2\% & 72.8\%  \\
1.1 & Creature & 398 & 9,009 & 143,996 & 130,409 & \bf 90.6\% & 61.4\% & 73.1\%  \\
1.1.1 & Domesticated animal & 123 & 2,316 & 50,036 & 41,978 & \bf 83.9\% & 63.4\% & 75.6\%  \\
1.1.2 & Vertebrate & 337 & 7,692 & 126,913 & 112,828 & \bf 88.9\% & 61.3\% & 73.2\%  \\
1.1.2.1 & Mammalian & 218 & 4,665 & 89,004 & 76,351 & \bf 85.8\% & 61.4\% & 73.5\%  \\
1.1.2.1.1 & Primate & 20 & 475 & 9,333 & 5,301 & \bf 56.8\% & 58.9\% & 70.4\%  \\
1.1.2.1.2 & Hoofed mammal & 17 & 419 & 6,206 & 2,751 & 44.3\% & 58.4\% & 71.6\%  \\
1.1.2.1.3 & Feline & 13 & 319 & 3,895 & 1,998 & \bf 51.3\% & 64.3\% & 75.9\%  \\
1.1.2.1.4 & Canine & 130 & 2,502 & 53,294 & 45,089 & \bf 84.6\% & 63.5\% & 75.7\%  \\
1.1.2.2 & Aquatic vertebrate & 16 & 366 & 5,355 & 2,383 & 44.5\% & 65.0\% & 75.6\%  \\
1.1.2.3 & Bird & 59 & 1,937 & 22,402 & 15,993 & \bf 71.4\% & 59.8\% & 71.3\%  \\
1.1.2.4 & Reptilian & 36 & 547 & 7,635 & 4,795 & \bf 62.8\% & 63.8\% & 75.2\%  \\
1.1.2.4.1 & Saurian & 11 & 188 & 2,416 & 1,050 & 43.5\% & 58.4\% & 71.1\%  \\
1.1.2.4.2 & Serpent & 17 & 223 & 3,202 & 1,700 & \bf 53.1\% & 67.0\% & 77.1\%  \\
1.1.3 & Invertebrate & 61 & 1,317 & 17,083 & 10,698 & \bf 62.6\% & 61.9\% & 72.3\%  \\
1.1.3.1 & Arthropod & 47 & 1,018 & 13,200 & 8,863 & \bf 67.1\% & 63.1\% & 73.5\%  \\
1.1.3.1.1 & Insect & 27 & 652 & 7,850 & 4,468 & \bf 56.9\% & 59.9\% & 70.5\%  \\
1.1.3.1.2 & Arachnoid & 9 & 189 & 2,824 & 1,476 & \bf  52.3\% & 69.7\% & 79.5\%  \\
1.1.3.1.3 & Crustacean & 9 & 137 & 2,035 & 955 & 46.9\% & 70.0\% & 80.1\%  \\
\cmidrule[0.25pt]{1-9}
2 & Artifact & 522 & 8,397 & 119,957 & 107,081 & \bf 89.3\% & 58.6\% & 70.2\%  \\
2.1 & Commodity & 63 & 906 & 16,092 & 5,411 & 33.6\% & 55.5\% & 68.6\%  \\
2.1.1 & Consumer Good & 62 & 896 & 15,923 & 5,205 & 32.7\% & 55.5\% & 68.6\%  \\
2.1.1.1 & Clothing & 49 & 670 & 12,010 & 4,660 & 38.8\% & 57.5\% & 70.8\%  \\
2.1.1.1.1 & Garment & 24 & 295 & 6,218 & 1,455 & 23.4\% & 56.4\% & 70.7\%  \\
2.1.1.2 & Durable & 13 & 226 & 3,913 & 331 & 8.5\% & 49.6\% & 61.8\%  \\
2.2 & Covering & 90 & 1,287 & 20,928 & 9,182 & 43.9\% & 59.4\% & 71.9\%  \\
2.2.1 & Protective covering & 27 & 407 & 6,021 & 766 & 12.7\% & 64.6\% & 75.7\%  \\
2.3 & Instrumentation & 353 & 5,963 & 80,638 & 55,364 & \bf  68.7\% & 58.0\% & 69.7\%  \\
2.3.1 & Container & 99 & 1,528 & 20,779 & 10,701 & \bf 51.5\% & 62.9\% & 73.5\%  \\
2.3.1.1 & Vessel & 23 & 261 & 4,515 & 1,373 & 30.4\% & 57.2\% & 67.9\%  \\
2.3.1.2 & Wheeled vehicle & 43 & 879 & 9,288 & 5,445 & \bf 58.6\% & 70.4\% & 80.0\%  \\
2.3.1.2.1 & Self-propelled vehicle & 31 & 627 & 6,761 & 3,336 & 49.3\% & 69.5\% & 79.7\%  \\
2.3.1.2.1.1 & Motor vehicle & 22 & 400 & 4,654 & 2,198 & 47.2\% & 67.6\% & 79.3\%  \\
2.3.2 & Transport & 71 & 1,558 & 17,929 & 10,643 & \bf 59.4\% & 64.5\% & 75.2\%  \\
2.3.2.1 & Vehicle & 66 & 1,439 & 16,790 & 9,439 & \bf 56.2\% & 64.3\% & 75.0\%  \\
2.3.2.1.1 & Air craft & 4 & 101 & 1,885 & 291 & 15.4\% & 50.7\% & 62.2\%  \\
2.3.2.1.2 & Water craft & 15 & 367 & 4,400 & 1,854 & 42.1\% & 59.5\% & 72.0\%  \\
2.3.3 & Device & 125 & 1,901 & 24,436 & 8,235 & 33.7\% & 57.5\% & 68.7\%  \\
2.3.3.1 & Instrument & 28 & 374 & 4,999 & 1,330 & 26.6\% & 57.6\% & 68.7\%  \\
2.3.3.1.1 & Measuring instrument & 12 & 202 & 2,605 & 716 & 27.5\% & 57.5\% & 67.4\%  \\
2.3.3.1.2 & Weapon & 7 & 69 & 914 & 150 & 16.4\% & 63.6\% & 72.2\%  \\
2.3.3.2 & Machine & 14 & 223 & 2,527 & 496 & 19.6\% & 69.7\% & 80.3\%  \\
2.3.3.3 & Mechanism & 12 & 219 & 2,814 & 45 & 1.6\% & 52.4\% & 63.8\%  \\
2.3.3.4 & Musical instrument & 26 & 427 & 4,756 & 1,835 & 38.6\% & 63.4\% & 74.1\%  \\
2.3.3.4.1 & Stringed instrument & 8 & 158 & 1,665 & 515 & 30.9\% & 61.7\% & 72.9\%  \\
2.3.3.4.2 & Wind instrument & 12 & 188 & 2,080 & 573 & 27.5\% & 63.3\% & 73.8\%  \\
2.3.4 & Equipment & 37 & 738 & 11,470 & 2,379 & 20.7\% & 50.2\% & 63.6\%  \\
2.3.4.1 & Electronic equipment & 13 & 178 & 3,122 & 394 & 12.6\% & 52.0\% & 64.9\%  \\
2.3.4.2 & Game equipment & 13 & 321 & 3,983 & 763 & 19.2\% & 56.3\% & 67.7\%  \\
2.3.5 & Furnishing & 25 & 447 & 7,554 & 1,774 & 23.5\% & 57.2\% & 69.6\%  \\
2.3.6 & Implement & 38 & 409 & 7,452 & 1,657 & 22.2\% & 57.2\% & 69.0\%  \\
2.4 & Structure & 57 & 1,035 & 12,799 & 5,349 & 41.8\% & 62.3\% & 72.1\%  \\
2.4.1 & Building & 14 & 293 & 3,428 & 663 & 19.3\% & 66.0\% & 76.5\%  \\
\cmidrule[0.25pt]{1-9}
3 & Geological formation & 10 & 139 & 3,631 & 1,439 & 39.6\% & 49.4\% & 61.2\%  \\
3.1 & Natural elevation & 5 & 65 & 1,705 & 219 & 12.8\% & 47.6\% & 60.1\%  \\
4 & Natural object & 17 & 379 & 5,734 & 1,700 & 29.6\% & 52.8\% & 63.4\%  \\
4.1 & Plant & 16 & 363 & 5,207 & 1,700 & 32.6\% & 53.7\% & 63.9\%  \\
4.1.1 & Fruit & 16 & 363 & 5,207 & 1,700 & 32.6\% & 53.7\% & 63.9\%  \\
4.1.1.1 & Edible fruit & 10 & 233 & 3,564 & 819 & 23.0\% & 49.7\% & 60.5\%  \\
5 & Fungus & 7 & 226 & 2,307 & 544 & 23.6\% & 56.1\% & 66.4\%  \\
6 & Nutrition & 10 & 157 & 3,017 & 528 & 17.5\% & 54.8\% & 64.1\%  \\
7 & Vegetable & 13 & 278 & 4,368 & 1,230 & 28.2\% & 56.5\% & 67.7\%  \\
8 & Beverage & 4 & 40 & 1,226 & 165 & 13.5\% & 64.4\% & 74.3\%  \\

\cmidrule[1pt]{1-9}
\end{tabular}
\label{tbl:main_pgd_cw_long}
\end{table}

\clearpage
\begin{table}[htbp!] 
\centering
\caption{For the adversarial examples that achieved model-to-model transferability and that have been created with 
\textbf{PGD}, intra-collection misclassifications and misclassifications into the top-\{3,5\} prediction classes in the target models are provided. The results for the adversarial examples are grouped into collections according to the classes of their source image origins.}
\scriptsize
\begin{tabular}{llccccccc}
\cmidrule[0.25pt]{1-9}
\multirow{4}{*}{\shortstack{Hierarchy}} & 
\multirow{4}{*}{\shortstack{Collection}} & 
\multirow{4}{*}{\shortstack{Classes\\in collection}} & 
\multirow{4}{*}{\shortstack{Source\\images\\in collection}} & 
\multirow{4}{*}{\shortstack{Adversarial\\examples\\originating\\from collection}} & 
\multicolumn{2}{c}{\multirow{3}{*}{\shortstack{Intra-collection\\misclassifications}}} &
\multicolumn{2}{c}{\multirow{3}{*}{\shortstack{Misclassification\\into top-K\\classes}}} \\
~ & ~ & ~ \\
~ & ~ & ~ \\
\cmidrule[0.25pt]{6-9}
~ & ~ & ~ & ~ & ~ &  Count & \%  & Top-3 & Top-5 \\
\cmidrule[0.25pt]{1-9}
~ & All & 1000 & 19,025 & 173,549 & 173,549 & 100.0\% & 59.5\% & 71.5\%  \\
\cmidrule[0.25pt]{1-9}
1 & Organism & 410 & 9,390 & 84,734 & 75,882 & \bf 89.6\% & 62.0\% & 74.0\%  \\
1.1 & Creature & 398 & 9,009 & 82,599 & 74,498 & \bf 90.2\% & 62.3\% & 74.2\%  \\
1.1.1 & Domesticated animal & 123 & 2,316 & 28,385 & 23,898 & \bf 84.2\% & 64.6\% & 77.2\%  \\
1.1.2 & Vertebrate & 337 & 7,692 & 72,329 & 64,258 & \bf 88.8\% & 62.3\% & 74.5\%  \\
1.1.2.1 & Mammalian & 218 & 4,665 & 50,125 & 43,705 & \bf 87.2\% & 62.9\% & 75.5\%  \\
1.1.2.1.1 & Primate & 20 & 475 & 5,123 & 2,999 & \bf 58.5\% & 60.4\% & 72.5\%  \\
1.1.2.1.2 & Hoofed mammal & 17 & 419 & 3,460 & 1,541 & 44.5\% & 60.2\% & 74.0\%  \\
1.1.2.1.3 & Feline & 13 & 319 & 2,346 & 1,262 & \bf 53.8\% & 65.9\% & 78.5\%  \\
1.1.2.1.4 & Canine & 130 & 2,502 & 30,094 & 25,784 & \bf 85.7\% & 64.8\% & 77.5\%  \\
1.1.2.2 & Aquatic vertebrate & 16 & 366 & 3,273 & 1,426 & 43.6\% & 64.7\% & 75.4\%  \\
1.1.2.3 & Bird & 59 & 1,937 & 12,878 & 9,013 & \bf 70.0\% & 60.3\% & 71.4\%  \\
1.1.2.4 & Reptilian & 36 & 547 & 4,549 & 2,829 & \bf 62.2\% & 62.7\% & 75.2\%  \\
1.1.2.4.1 & Saurian & 11 & 188 & 1,449 & 610 & 42.1\% & 56.5\% & 70.2\%  \\
1.1.2.4.2 & Serpent & 17 & 223 & 1,931 & 1,013 & \bf 52.5\% & 66.0\% & 77.3\%  \\
1.1.3 & Invertebrate & 61 & 1,317 & 10,270 & 6,329 & \bf 61.6\% & 62.0\% & 72.5\%  \\
1.1.3.1 & Arthropod & 47 & 1,018 & 7,893 & 5,200 & \bf 65.9\% & 63.1\% & 73.7\%  \\
1.1.3.1.1 & Insect & 27 & 652 & 4,650 & 2,566 & \bf 55.2\% & 59.7\% & 70.5\%  \\
1.1.3.1.2 & Arachnoid & 9 & 189 & 1,700 & 932 & \bf 54.8\% & 70.0\% & 80.1\%  \\
1.1.3.1.3 & Crustacean & 9 & 137 & 1,247 & 571 & 45.8\% & 70.2\% & 80.5\%  \\
\cmidrule[0.25pt]{1-9}
2 & Artifact & 522 & 8,397 & 75,248 & 67,853 & \bf 90.2\% & 57.7\% & 70.0\%  \\
2.1 & Commodity & 63 & 906 & 10,204 & 3,428 & 33.6\% & 54.7\% & 68.5\%  \\
2.1.1 & Consumer Good & 62 & 896 & 10,107 & 3,290 & 32.6\% & 54.7\% & 68.4\%  \\
2.1.1.1 & Clothing & 49 & 670 & 7,515 & 2,984 & 39.7\% & 56.8\% & 71.0\%  \\
2.1.1.1.1 & Garment & 24 & 295 & 3,877 & 928 & 23.9\% & 55.4\% & 70.6\%  \\
2.1.1.2 & Durable & 13 & 226 & 2,592 & 187 & 7.2\% & 48.3\% & 60.8\%  \\
2.2 & Covering & 90 & 1,287 & 13,113 & 5,846 & 44.6\% & 58.3\% & 71.6\%  \\
2.2.1 & Protective covering & 27 & 407 & 3,793 & 511 & 13.5\% & 63.2\% & 74.7\%  \\
2.3 & Instrumentation & 353 & 5,963 & 50,597 & 34,722 & \bf 68.6\% & 57.1\% & 69.4\%  \\
2.3.1 & Container & 99 & 1,528 & 12,966 & 6,622 & \bf 51.1\% & 61.8\% & 72.9\%  \\
2.3.1.1 & Vessel & 23 & 261 & 2,789 & 804 & 28.8\% & 55.3\% & 66.0\%  \\
2.3.1.2 & Wheeled vehicle & 43 & 879 & 5,791 & 3,403 & \bf 58.8\% & 70.1\% & 80.2\%  \\
2.3.1.2.1 & Self-propelled vehicle & 31 & 627 & 4,262 & 2,126 & 49.9\% & 69.4\% & 80.2\%  \\
2.3.1.2.1.1 & Motor vehicle & 22 & 400 & 2,953 & 1,406 & 47.6\% & 67.7\% & 80.1\%  \\
2.3.2 & Transport & 71 & 1,558 & 11,340 & 6,725 & \bf 59.3\% & 63.8\% & 75.1\%  \\
2.3.2.1 & Vehicle & 66 & 1,439 & 10,604 & 5,946 & \bf 56.1\% & 63.6\% & 74.9\%  \\
2.3.2.1.1 & Air craft & 4 & 101 & 1,180 & 193 & 16.4\% & 49.0\% & 61.6\%  \\
2.3.2.1.2 & Water craft & 15 & 367 & 2,845 & 1,167 & 41.0\% & 58.9\% & 71.7\%  \\
2.3.3 & Device & 125 & 1,901 & 15,419 & 5,212 & 33.8\% & 56.7\% & 68.8\%  \\
2.3.3.1 & Instrument & 28 & 374 & 3,088 & 836 & 27.1\% & 58.0\% & 69.3\%  \\
2.3.3.1.1 & Measuring instrument & 12 & 202 & 1,624 & 468 & 28.8\% & 57.3\% & 67.4\%  \\
2.3.3.1.2 & Weapon & 7 & 69 & 527 & 86 & 16.3\% & 66.6\% & 74.8\%  \\
2.3.3.2 & Machine & 14 & 223 & 1,690 & 293 & 17.3\% & 67.6\% & 79.2\%  \\
2.3.3.3 & Mechanism & 12 & 219 & 1,809 & 29 & 1.6\% & 51.1\% & 63.1\%  \\
2.3.3.4 & Musical instrument & 26 & 427 & 2,912 & 1,155 & 39.7\% & 62.7\% & 75.2\%  \\
2.3.3.4.1 & Stringed instrument & 8 & 158 & 1,015 & 324 & 31.9\% & 61.1\% & 74.3\%  \\
2.3.3.4.2 & Wind instrument & 12 & 188 & 1,283 & 374 & 29.2\% & 63.0\% & 74.8\%  \\
2.3.4 & Equipment & 37 & 738 & 7,257 & 1,555 & 21.4\% & 49.4\% & 64.0\%  \\
2.3.4.1 & Electronic equipment & 13 & 178 & 1,947 & 251 & 12.9\% & 49.3\% & 63.5\%  \\
2.3.4.2 & Game equipment & 13 & 321 & 2,538 & 510 & 20.1\% & 57.1\% & 69.3\%  \\
2.3.5 & Furnishing & 25 & 447 & 4,697 & 1,067 & 22.7\% & 55.5\% & 68.4\%  \\
2.3.6 & Implement & 38 & 409 & 4,544 & 1,013 & 22.3\% & 56.8\% & 69.5\%  \\
2.4 & Structure & 57 & 1,035 & 7,998 & 3,404 & 42.6\% & 62.5\% & 72.8\%  \\
2.4.1 & Building & 14 & 293 & 2,137 & 431 & 20.2\% & 65.2\% & 76.5\%  \\
\cmidrule[0.25pt]{1-9}
3 & Geological formation & 10 & 139 & 2,250 & 860 & 38.2\% & 46.8\% & 59.8\%  \\
3.1 & Natural elevation & 5 & 65 & 1,080 & 123 & 11.4\% & 44.1\% & 58.2\%  \\
4 & Natural object & 17 & 379 & 3,590 & 1,105 & 30.8\% & 52.2\% & 64.3\%  \\
4.1 & Plant & 16 & 363 & 3,238 & 1,105 & 34.1\% & 53.6\% & 64.8\%  \\
4.1.1 & Fruit & 16 & 363 & 3,238 & 1,105 & 34.1\% & 53.6\% & 64.8\%  \\
4.1.1.1 & Edible fruit & 10 & 233 & 2,250 & 550 & 24.4\% & 49.0\% & 61.1\%  \\
5 & Fungus & 7 & 226 & 1,320 & 295 & 22.3\% & 55.4\% & 65.9\%  \\
6 & Nutrition & 10 & 157 & 1,895 & 340 & 17.9\% & 53.9\% & 63.9\%  \\
7 & Vegetable & 13 & 278 & 2,814 & 772 & 27.4\% & 56.1\% & 68.0\%  \\
8 & Beverage & 4 & 40 & 767 & 93 & 12.1\% & 61.4\% & 71.7\%  \\
\cmidrule[1pt]{1-9}
\end{tabular}
\label{tbl:main_pgd_long}
\end{table}

\clearpage
\begin{table}[htbp!] 
\centering
\caption{For the adversarial examples that achieved model-to-model transferability and that have been created with 
\textbf{CW}, intra-collection misclassifications and misclassifications into the top-\{3,5\} prediction classes in the target models are provided. The results for the adversarial examples are grouped into collections according to the classes of their source image origins.}
\scriptsize
\begin{tabular}{llccccccc}
\cmidrule[0.25pt]{1-9}
\multirow{4}{*}{\shortstack{Hierarchy}} & 
\multirow{4}{*}{\shortstack{Collection}} & 
\multirow{4}{*}{\shortstack{Classes\\in collection}} & 
\multirow{4}{*}{\shortstack{Source\\images\\in collection}} & 
\multirow{4}{*}{\shortstack{Adversarial\\examples\\originating\\from collection}} & 
\multicolumn{2}{c}{\multirow{3}{*}{\shortstack{Intra-collection\\misclassifications}}} &
\multicolumn{2}{c}{\multirow{3}{*}{\shortstack{Misclassification\\into top-K\\classes}}} \\
~ & ~ & ~ \\
~ & ~ & ~ \\
\cmidrule[0.25pt]{6-9}
~ & ~ & ~ & ~ & ~ &  Count & \%  & Top-3 & Top-5 \\
\cmidrule[0.25pt]{1-9}
~ & All & 1000 & 19,025 & 115,695 & 115,695 & 100.0\% & 59.8\% & 70.5\%  \\
\cmidrule[0.25pt]{1-9}
1 & Organism & 410 & 9,390 & 62,887 & 56,983 & \bf 90.6\% & 60.1\% & 71.3\%  \\
1.1 & Creature & 398 & 9,009 & 61,397 & 55,911 & \bf 91.1\% & 60.2\% & 71.5\%  \\
1.1.1 & Domesticated animal & 123 & 2,316 & 21,651 & 18,080 & \bf 83.5\% & 61.8\% & 73.5\%  \\
1.1.2 & Vertebrate & 337 & 7,692 & 54,584 & 48,570 & \bf 89.0\% & 60.0\% & 71.4\%  \\
1.1.2.1 & Mammalian & 218 & 4,665 & 38,879 & 32,646 & \bf 84.0\% & 59.6\% & 71.0\%  \\
1.1.2.1.1 & Primate & 20 & 475 & 4,210 & 2,302 & \bf 54.7\% & 57.1\% & 67.8\%  \\
1.1.2.1.2 & Hoofed mammal & 17 & 419 & 2,746 & 1,210 & 44.1\% & 56.2\% & 68.6\%  \\
1.1.2.1.3 & Feline & 13 & 319 & 1,549 & 736 & 47.5\% & 61.9\% & 72.0\%  \\
1.1.2.1.4 & Canine & 130 & 2,502 & 23,200 & 19,305 & \bf 83.2\% & 61.8\% & 73.5\%  \\
1.1.2.2 & Aquatic vertebrate & 16 & 366 & 2,082 & 957 & 46.0\% & 65.6\% & 75.9\%  \\
1.1.2.3 & Bird & 59 & 1,937 & 9,524 & 6,980 & \bf 73.3\% & 59.2\% & 71.1\%  \\
1.1.2.4 & Reptilian & 36 & 547 & 3,086 & 1,966 & \bf 63.7\% & 65.4\% & 75.1\%  \\
1.1.2.4.1 & Saurian & 11 & 188 & 967 & 440 & 45.5\% & 61.4\% & 72.4\%  \\
1.1.2.4.2 & Serpent & 17 & 223 & 1,271 & 687 & \bf 54.1\% & 68.5\% & 76.8\%  \\
1.1.3 & Invertebrate & 61 & 1,317 & 6,813 & 4,369 & \bf 64.1\% & 61.8\% & 72.1\%  \\
1.1.3.1 & Arthropod & 47 & 1,018 & 5,307 & 3,663 & \bf 69.0\% & 63.0\% & 73.3\%  \\
1.1.3.1.1 & Insect & 27 & 652 & 3,200 & 1,902 & \bf 59.4\% & 60.2\% & 70.5\%  \\
1.1.3.1.2 & Arachnoid & 9 & 189 & 1,124 & 544 & 48.4\% & 69.3\% & 78.6\%  \\
1.1.3.1.3 & Crustacean & 9 & 137 & 788 & 384 & 48.7\% & 69.7\% & 79.4\%  \\
\cmidrule[0.25pt]{1-9}
2 & Artifact & 522 & 8,397 & 44,709 & 39,228 & \bf 87.7\% & 60.1\% & 70.5\%  \\
2.1 & Commodity & 63 & 906 & 5,888 & 1,983 & 33.7\% & 56.9\% & 68.8\%  \\
2.1.1 & Consumer Good & 62 & 896 & 5,816 & 1,915 & 32.9\% & 57.1\% & 68.9\%  \\
2.1.1.1 & Clothing & 49 & 670 & 4,495 & 1,676 & 37.3\% & 58.5\% & 70.4\%  \\
2.1.1.1.1 & Garment & 24 & 295 & 2,341 & 527 & 22.5\% & 58.1\% & 70.9\%  \\
2.1.1.2 & Durable & 13 & 226 & 1,321 & 144 & 10.9\% & 52.2\% & 63.7\%  \\
2.2 & Covering & 90 & 1,287 & 7,815 & 3,336 & 42.7\% & 61.1\% & 72.4\%  \\
2.2.1 & Protective covering & 27 & 407 & 2,228 & 255 & 11.4\% & 66.9\% & 77.4\%  \\
2.3 & Instrumentation & 353 & 5,963 & 30,041 & 20,642 & \bf 68.7\% & 59.7\% & 70.1\%  \\
2.3.1 & Container & 99 & 1,528 & 7,813 & 4,079 & \bf 52.2\% & 64.6\% & 74.4\%  \\
2.3.1.1 & Vessel & 23 & 261 & 1,726 & 569 & 33.0\% & 60.4\% & 71.0\%  \\
2.3.1.2 & Wheeled vehicle & 43 & 879 & 3,497 & 2,042 & \bf 58.4\% & 71.1\% & 79.5\%  \\
2.3.1.2.1 & Self-propelled vehicle & 31 & 627 & 2,499 & 1,210 & 48.4\% & 69.8\% & 78.8\%  \\
2.3.1.2.1.1 & Motor vehicle & 22 & 400 & 1,701 & 792 & 46.6\% & 67.4\% & 78.0\%  \\
2.3.2 & Transport & 71 & 1,558 & 6,589 & 3,918 & \bf 59.5\% & 65.7\% & 75.4\%  \\
2.3.2.1 & Vehicle & 66 & 1,439 & 6,186 & 3,493 & \bf 56.5\% & 65.5\% & 75.2\%  \\
2.3.2.1.1 & Air craft & 4 & 101 & 705 & 98 & 13.9\% & 53.5\% & 63.1\%  \\
2.3.2.1.2 & Water craft & 15 & 367 & 1,555 & 687 & 44.2\% & 60.6\% & 72.7\%  \\
2.3.3 & Device & 125 & 1,901 & 9,017 & 3,023 & 33.5\% & 58.9\% & 68.6\%  \\
2.3.3.1 & Instrument & 28 & 374 & 1,911 & 494 & 25.9\% & 56.9\% & 67.8\%  \\
2.3.3.1.1 & Measuring instrument & 12 & 202 & 981 & 248 & 25.3\% & 57.8\% & 67.4\%  \\
2.3.3.1.2 & Weapon & 7 & 69 & 387 & 64 & 16.5\% & 59.4\% & 68.7\%  \\
2.3.3.2 & Machine & 14 & 223 & 837 & 203 & 24.3\% & 74.0\% & 82.6\%  \\
2.3.3.3 & Mechanism & 12 & 219 & 1,005 & 16 & 1.6\% & 54.7\% & 65.1\%  \\
2.3.3.4 & Musical instrument & 26 & 427 & 1,844 & 680 & 36.9\% & 64.5\% & 72.4\%  \\
2.3.3.4.1 & Stringed instrument & 8 & 158 & 650 & 191 & 29.4\% & 62.8\% & 70.8\%  \\
2.3.3.4.2 & Wind instrument & 12 & 188 & 797 & 199 & 25.0\% & 63.7\% & 72.3\%  \\
2.3.4 & Equipment & 37 & 738 & 4,213 & 824 & 19.6\% & 51.7\% & 62.8\%  \\
2.3.4.1 & Electronic equipment & 13 & 178 & 1,175 & 143 & 12.2\% & 56.4\% & 67.2\%  \\
2.3.4.2 & Game equipment & 13 & 321 & 1,445 & 253 & 17.5\% & 54.9\% & 64.8\%  \\
2.3.5 & Furnishing & 25 & 447 & 2,857 & 707 & 24.7\% & 60.0\% & 71.6\%  \\
2.3.6 & Implement & 38 & 409 & 2,908 & 644 & 22.1\% & 57.9\% & 68.3\%  \\
2.4 & Structure & 57 & 1,035 & 4,801 & 1,945 & 40.5\% & 62.2\% & 71.0\%  \\
2.4.1 & Building & 14 & 293 & 1,291 & 232 & 18.0\% & 67.4\% & 76.4\%  \\
3 & Geological formation & 10 & 139 & 1,381 & 579 & 41.9\% & 53.8\% & 63.4\%  \\
\cmidrule[0.25pt]{1-9}
3.1 & Natural elevation & 5 & 65 & 625 & 96 & 15.4\% & 53.8\% & 63.2\%  \\
4 & Natural object & 17 & 379 & 2,144 & 595 & 27.8\% & 53.7\% & 61.9\%  \\
4.1 & Plant & 16 & 363 & 1,969 & 595 & 30.2\% & 53.9\% & 62.3\%  \\
4.1.1 & Fruit & 16 & 363 & 1,969 & 595 & 30.2\% & 53.9\% & 62.3\%  \\
4.1.1.1 & Edible fruit & 10 & 233 & 1,314 & 269 & 20.5\% & 50.9\% & 59.5\%  \\
5 & Fungus & 7 & 226 & 987 & 249 & 25.2\% & 57.1\% & 67.2\%  \\
6 & Nutrition & 10 & 157 & 1,122 & 188 & 16.8\% & 56.3\% & 64.4\%  \\
7 & Vegetable & 13 & 278 & 1,554 & 458 & 29.5\% & 57.3\% & 67.3\%  \\
8 & Beverage & 4 & 40 & 459 & 72 & 15.7\% & 69.5\% & 78.6\%  \\

\cmidrule[1pt]{1-9}
\end{tabular}
\label{tbl:main_cw_long}
\end{table}

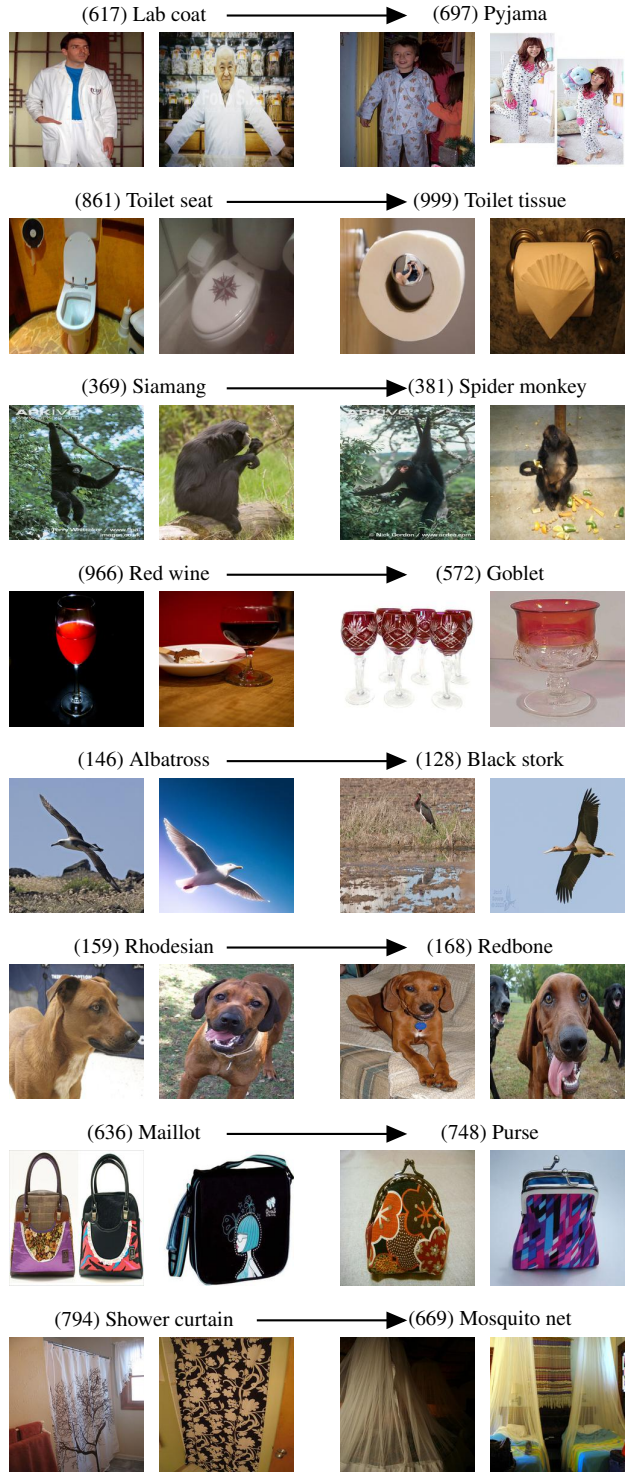
\begin{figure}[hbtp!]
\centering
\begin{tikzpicture}[thick,scale=0.8, every node/.style={scale=0.8}]

\def\xpos{2}
\def\ypos{-1}
\node[] at (\xpos-2.875, \ypos+0.8)  {(617) Lab coat};
\node[] at (\xpos+2.875, \ypos+0.8)  {(697) Pyjama};
\node[inner sep=0pt] (vid2) at (\xpos -4, \ypos-0.6)
    {\includegraphics[width=.16\textwidth]{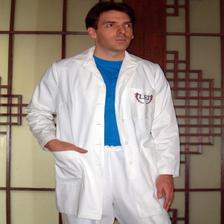}};
\node[inner sep=0pt] (vid2) at (\xpos -1.5, \ypos-0.6)
    {\includegraphics[width=.16\textwidth]{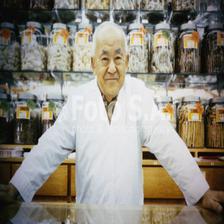}};
\node[inner sep=0pt] (vid2) at (\xpos +1.5, \ypos-0.6)
    {\includegraphics[width=.16\textwidth]{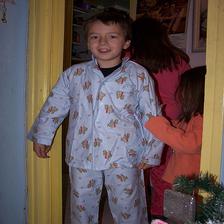}};
\node[inner sep=0pt] (vid2) at (\xpos +4, \ypos-0.6)
    {\includegraphics[width=.16\textwidth]{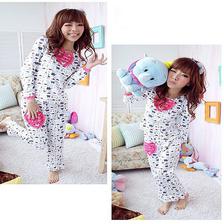}};
\draw[>=triangle 45, ->] (\xpos-1.5, \ypos+0.8) -- (\xpos+1.5, \ypos+0.8);
\def\ypos{-4.1}
\node[] at (\xpos-2.875, \ypos+0.8)  {(861) Toilet seat};
\node[] at (\xpos+2.875, \ypos+0.8)  {(999) Toilet tissue};
\node[inner sep=0pt] (vid2) at (\xpos -4, \ypos-0.6)
    {\includegraphics[width=.16\textwidth]{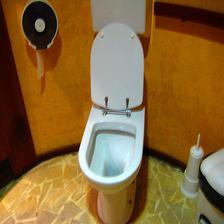}};
\node[inner sep=0pt] (vid2) at (\xpos -1.5, \ypos-0.6)
    {\includegraphics[width=.16\textwidth]{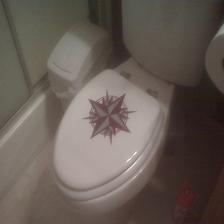}};
\node[inner sep=0pt] (vid2) at (\xpos +1.5, \ypos-0.6)
    {\includegraphics[width=.16\textwidth]{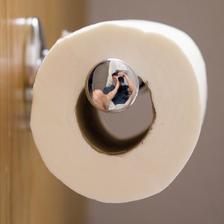}};
\node[inner sep=0pt] (vid2) at (\xpos +4, \ypos-0.6)
    {\includegraphics[width=.16\textwidth]{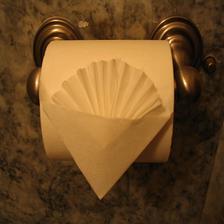}};
\draw[>=triangle 45, ->] (\xpos-1.5, \ypos+0.8) -- (\xpos+1.5, \ypos+0.8);
\def\ypos{-7.2}
\node[] at (\xpos-2.875, \ypos+0.8)  {(369) Siamang};
\node[] at (\xpos+3, \ypos+0.8)  {(381) Spider monkey};
\node[inner sep=0pt] (vid2) at (\xpos -4, \ypos-0.6)
    {\includegraphics[width=.16\textwidth]{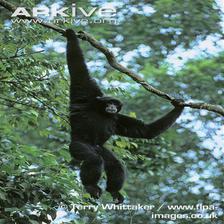}};
\node[inner sep=0pt] (vid2) at (\xpos -1.5, \ypos-0.6)
    {\includegraphics[width=.16\textwidth]{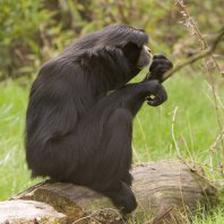}};
\node[inner sep=0pt] (vid2) at (\xpos +1.5, \ypos-0.6)
    {\includegraphics[width=.16\textwidth]{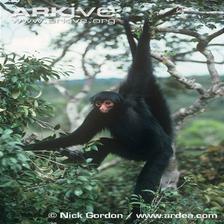}};
\node[inner sep=0pt] (vid2) at (\xpos +4, \ypos-0.6)
    {\includegraphics[width=.16\textwidth]{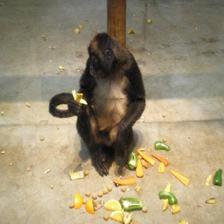}};
\draw[>=triangle 45, ->] (\xpos-1.5, \ypos+0.8) -- (\xpos+1.5, \ypos+0.8);
\def\ypos{-10.3}
\node[] at (\xpos-2.875, \ypos+0.8)  {(966) Red wine};
\node[] at (\xpos+2.875, \ypos+0.8)  {(572) Goblet};
\node[inner sep=0pt] (vid2) at (\xpos -4, \ypos-0.6)
    {\includegraphics[width=.16\textwidth]{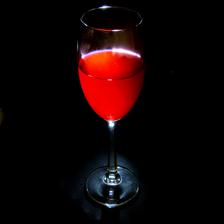}};
\node[inner sep=0pt] (vid2) at (\xpos -1.5, \ypos-0.6)
    {\includegraphics[width=.16\textwidth]{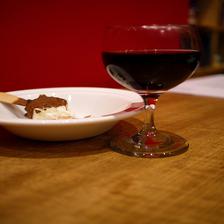}};
\node[inner sep=0pt] (vid2) at (\xpos +1.5, \ypos-0.6)
    {\includegraphics[width=.16\textwidth]{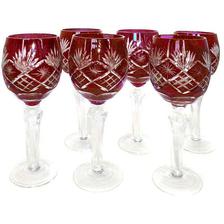}};
\node[inner sep=0pt] (vid2) at (\xpos +4, \ypos-0.6)
    {\includegraphics[width=.16\textwidth]{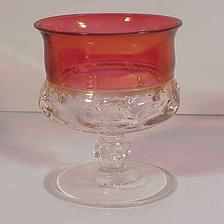}};
\draw[>=triangle 45, ->] (\xpos-1.5, \ypos+0.8) -- (\xpos+1.5, \ypos+0.8);
\def\ypos{-13.4}
\node[] at (\xpos-2.875, \ypos+0.8)  {(146) Albatross};
\node[] at (\xpos+2.875, \ypos+0.8)  {(128) Black stork};
\node[inner sep=0pt] (vid2) at (\xpos -4, \ypos-0.6)
    {\includegraphics[width=.16\textwidth]{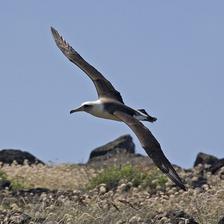}};
\node[inner sep=0pt] (vid2) at (\xpos -1.5, \ypos-0.6)
    {\includegraphics[width=.16\textwidth]{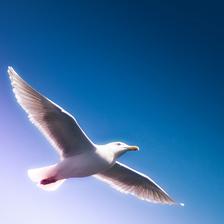}};
\node[inner sep=0pt] (vid2) at (\xpos +1.5, \ypos-0.6)
    {\includegraphics[width=.16\textwidth]{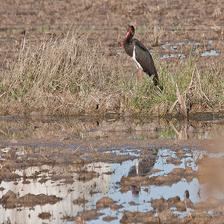}};
\node[inner sep=0pt] (vid2) at (\xpos +4, \ypos-0.6)
    {\includegraphics[width=.16\textwidth]{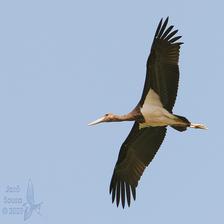}};
\draw[>=triangle 45, ->] (\xpos-1.5, \ypos+0.8) -- (\xpos+1.5, \ypos+0.8);
\def\ypos{-16.5}
\node[] at (\xpos-2.875, \ypos+0.8)  {(159) Rhodesian};
\node[] at (\xpos+2.875, \ypos+0.8)  {(168) Redbone};
\node[inner sep=0pt] (vid2) at (\xpos -4, \ypos-0.6)
    {\includegraphics[width=.16\textwidth]{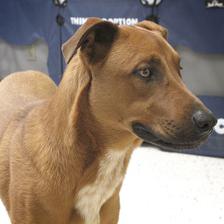}};
\node[inner sep=0pt] (vid2) at (\xpos -1.5, \ypos-0.6)
    {\includegraphics[width=.16\textwidth]{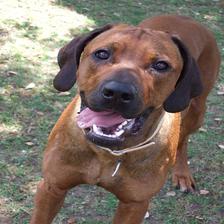}};
\node[inner sep=0pt] (vid2) at (\xpos +1.5, \ypos-0.6)
    {\includegraphics[width=.16\textwidth]{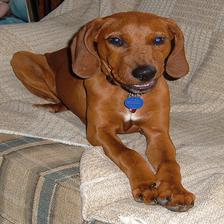}};
\node[inner sep=0pt] (vid2) at (\xpos +4, \ypos-0.6)
    {\includegraphics[width=.16\textwidth]{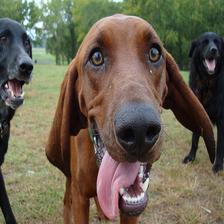}};
\draw[>=triangle 45, ->] (\xpos-1.5, \ypos+0.8) -- (\xpos+1.5, \ypos+0.8);
\def\ypos{-19.6}
\node[] at (\xpos-2.875, \ypos+0.8)  {(636) Maillot};
\node[] at (\xpos+2.875, \ypos+0.8)  {(748) Purse};
\node[inner sep=0pt] (vid2) at (\xpos -4, \ypos-0.6)
    {\includegraphics[width=.16\textwidth]{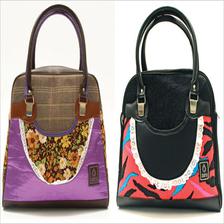}};
\node[inner sep=0pt] (vid2) at (\xpos -1.5, \ypos-0.6)
    {\includegraphics[width=.16\textwidth]{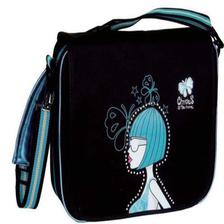}};
\node[inner sep=0pt] (vid2) at (\xpos +1.5, \ypos-0.6)
    {\includegraphics[width=.16\textwidth]{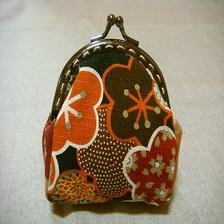}};
\node[inner sep=0pt] (vid2) at (\xpos +4, \ypos-0.6)
    {\includegraphics[width=.16\textwidth]{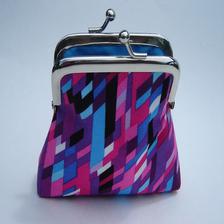}};
\draw[>=triangle 45, ->] (\xpos-1.5, \ypos+0.8) -- (\xpos+1.5, \ypos+0.8);
\def\ypos{-22.7}
\node[] at (\xpos-2.875, \ypos+0.8)  {(794) Shower curtain};
\node[] at (\xpos+2.875, \ypos+0.8)  {(669) Mosquito net};
\node[inner sep=0pt] (vid2) at (\xpos -4, \ypos-0.6)
    {\includegraphics[width=.16\textwidth]{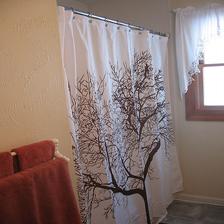}};
\node[inner sep=0pt] (vid2) at (\xpos -1.5, \ypos-0.6)
    {\includegraphics[width=.16\textwidth]{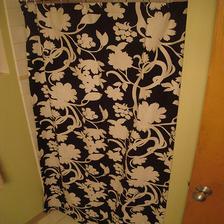}};
\node[inner sep=0pt] (vid2) at (\xpos +1.5, \ypos-0.6)
    {\includegraphics[width=.16\textwidth]{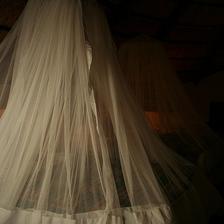}};
\node[inner sep=0pt] (vid2) at (\xpos +4, \ypos-0.6)
    {\includegraphics[width=.16\textwidth]{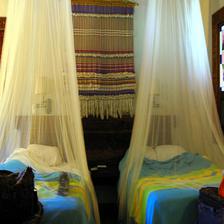}};
\draw[>=triangle 45, ->] (\xpos-1, \ypos+0.8) -- (\xpos+1.5, \ypos+0.8);
\end{tikzpicture}
\caption{Adversarial examples on the left are misclassified into the classes on the right by multiple models used in this study. The classes given on the right often lie in the top-5 predictions for the genuine source image counterparts of those adversarial examples.}
\label{fig:imagenet-similar-classes_sup}
\end{figure}

\clearpage
\begin{table}[htbp!] 
\centering
\caption{For the adversarial examples that achieved model-to-model transferability and that have been created with \textbf{PGD} and \textbf{CW}, intra-collection misclassifications and misclassifications into the top-\{3,5\} prediction classes in the target models are provided for each model employed in this study ($1$st column). The results for the adversarial examples are grouped into collections according to the classes of their source image origins. The results are provided for a number of collections that lie under the \textbf{Organism} sub-tree.}
\scriptsize
\begin{tabular}{lllccccccc}
\cmidrule[0.25pt]{1-10}
\parbox[t]{2mm}{\multirow{4}{*}{\rotatebox[origin=c]{90}{Model}}} & 
\multirow{4}{*}{\shortstack{Hierarchy}} & 
\multirow{4}{*}{\shortstack{Collection}} & 
\multirow{4}{*}{\shortstack{Classes\\in collection}} & 
\multirow{4}{*}{\shortstack{Source\\images\\in collection}} & 
\multirow{4}{*}{\shortstack{Adversarial\\examples\\originating\\from collection}} & 
\multicolumn{2}{c}{\multirow{3}{*}{\shortstack{Intra-collection\\misclassifications}}} &
\multicolumn{2}{c}{\multirow{3}{*}{\shortstack{Misclassification\\into top-K\\classes}}} \\
~ & ~ & ~ \\
~ & ~ & ~ \\
\cmidrule[0.25pt]{7-10}
~ & ~ & ~ & ~ & ~ &  ~ & Count & \%  & Top-3 & Top-5 \\
\cmidrule[0.25pt]{1-10}
\parbox[t]{2mm}{\multirow{9}{*}{\rotatebox[origin=c]{90}{AlexNet}}}  &  1 & Organism & 410 & 9,390 & 23,841 & 21,977 & \bf 92.2\% & 76.0\% & 86.8\%  \\
~ &  1.1.2.1.1 & Primate & 20 & 475 & 1,587 & 755 & 47.6\% & 78.3\% & 88.6\%  \\
~ &  1.1.2.1.2 & Hoofed mammal & 17 & 419 & 1,044 & 420 & 40.2\% & 70.1\% & 86.9\%  \\
~ &  1.1.2.1.3 & Feline & 13 & 319 & 781 & 354 & 45.3\% & 74.6\% & 89.2\%  \\
~ &  1.1.2.1.4 & Canine & 130 & 2,502 & 8,709 & 7,112 & \bf 81.7\% & 77.2\% & 87.7\%  \\
~ &  1.1.2.2 & Aquatic vertebrate & 16 & 366 & 721 & 313 & 43.4\% & 84.6\% & 92.8\%  \\
~ &  1.1.2.3 & Bird & 59 & 1,937 & 3,841 & 2,732 & \bf 71.1\% & 74.2\% & 85.1\%  \\
~ &  1.1.2.4 & Reptilian & 36 & 547 & 1,415 & 832 & \bf 58.8\% & 73.5\% & 86.1\%  \\
~ &  1.1.3 & Invertebrate & 61 & 1,317 & 2,620 & 1,740 & \bf 66.4\% & 75.6\% & 84.9\%  \\
\cmidrule[0.25pt]{1-10}
\parbox[t]{2mm}{\multirow{9}{*}{\rotatebox[origin=c]{90}{SqueezeNet}}}  &  1 & Organism & 410 & 9,390 & 41,266 & 36,530 & \bf 88.5\% & 62.5\% & 75.4\%  \\
~ &  1.1.2.1.1 & Primate & 20 & 475 & 2,589 & 1,235 & 47.7\% & 61.7\% & 73.7\%  \\
~ &  1.1.2.1.2 & Hoofed mammal & 17 & 419 & 1,909 & 699 & 36.6\% & 60.9\% & 74.5\%  \\
~ &  1.1.2.1.3 & Feline & 13 & 319 & 1,267 & 563 & 44.4\% & 62.7\% & 76.0\%  \\
~ &  1.1.2.1.4 & Canine & 130 & 2,502 & 13,931 & 11,172 & \bf 80.2\% & 63.2\% & 76.5\%  \\
~ &  1.1.2.2 & Aquatic vertebrate & 16 & 366 & 1,459 & 530 & 36.3\% & 66.0\% & 77.9\%  \\
~ &  1.1.2.3 & Bird & 59 & 1,937 & 6,850 & 4,476 & \bf 65.3\% & 61.3\% & 73.6\%  \\
~ &  1.1.2.4 & Reptilian & 36 & 547 & 2,349 & 1,348 & \bf 57.4\% & 65.7\% & 78.2\%  \\
~ &  1.1.3 & Invertebrate & 61 & 1,317 & 4,900 & 2,615 & \bf  53.4\% & 62.1\% & 74.8\%  \\
\cmidrule[0.25pt]{1-10}
\parbox[t]{2mm}{\multirow{9}{*}{\rotatebox[origin=c]{90}{VGG-16}}}  &  1 & Organism & 410 & 9,390 & 25,580 & 23,658 & \bf  92.5\% & 56.1\% & 68.7\%  \\
~ &  1.1.2.1.1 & Primate & 20 & 475 & 1,589 & 1,051 & \bf  66.1\% & 52.5\% & 63.6\%  \\
~ &  1.1.2.1.2 & Hoofed mammal & 17 & 419 & 999 & 511 & \bf 51.2\% & 53.6\% & 66.6\%  \\
~ &  1.1.2.1.3 & Feline & 13 & 319 & 570 & 332 & \bf 58.2\% & 62.3\% & 73.3\%  \\
~ &  1.1.2.1.4 & Canine & 130 & 2,502 & 9,241 & 7,901 & \bf  85.5\% & 58.3\% & 71.3\%  \\
~ &  1.1.2.2 & Aquatic vertebrate & 16 & 366 & 1,017 & 472 &  46.4\% & 59.1\% & 70.9\%  \\
~ &  1.1.2.3 & Bird & 59 & 1,937 & 4,085 & 3,200 & \bf 78.3\% & 55.0\% & 68.6\%  \\
~ &  1.1.2.4 & Reptilian & 36 & 547 & 1,096 & 784 & \bf 71.5\% & 62.0\% & 75.8\%  \\
~ &  1.1.3 & Invertebrate & 61 & 1,317 & 2,969 & 1,933 & \bf  65.1\% & 57.5\% & 67.7\%  \\
\cmidrule[0.25pt]{1-10}
\parbox[t]{2mm}{\multirow{9}{*}{\rotatebox[origin=c]{90}{DenseNet-121}}}  &  1 & Organism & 410 & 9,390 & 16,477 & 15,181 & \bf 92.1\% & 64.3\% & 75.3\%  \\
~ &  1.1.2.1.1 & Primate & 20 & 475 & 1,019 & 697 & \bf 68.4\% & 61.7\% & 73.7\%  \\
~ &  1.1.2.1.2 & Hoofed mammal & 17 & 419 & 650 & 335 & \bf 51.5\% & 59.5\% & 72.2\%  \\
~ &  1.1.2.1.3 & Feline & 13 & 319 & 248 & 161 & \bf 64.9\% & 73.0\% & 79.0\%  \\
~ &  1.1.2.1.4 & Canine & 130 & 2,502 & 6,150 & 5,596 & \bf 91.0\% & 67.6\% & 79.9\%  \\
~ &  1.1.2.2 & Aquatic vertebrate & 16 & 366 & 671 & 363 & \bf 54.1\% & 67.1\% & 76.5\%  \\
~ &  1.1.2.3 & Bird & 59 & 1,937 & 2,260 & 1,731 & \bf 76.6\% & 65.7\% & 75.6\%  \\
~ &  1.1.2.4 & Reptilian & 36 & 547 & 844 & 551 & \bf 65.3\% & 61.5\% & 70.5\%  \\
~ &  1.1.3 & Invertebrate & 61 & 1,317 & 1,963 & 1,343 & \bf 68.4\% & 63.3\% & 72.6\%  \\
\cmidrule[0.25pt]{1-10}
\parbox[t]{2mm}{\multirow{9}{*}{\rotatebox[origin=c]{90}{ResNet-50}}}  &  1 & Organism & 410 & 9,390 & 17,487 & 15,948 & 91.2\% & 59.6\% & 70.8\%  \\
~ &  1.1.2.1.1 & Primate & 20 & 475 & 1,232 & 695 & \bf 56.4\% & 50.0\% & 62.9\%  \\
~ &  1.1.2.1.2 & Hoofed mammal & 17 & 419 & 790 & 407 & \bf 51.5\% & 54.3\% & 67.8\%  \\
~ &  1.1.2.1.3 & Feline & 13 & 319 & 318 & 217 & \bf 68.2\% & 70.8\% & 76.7\%  \\
~ &  1.1.2.1.4 & Canine & 130 & 2,502 & 6,346 & 5,566 & \bf 87.7\% & 62.4\% & 74.2\%  \\
~ &  1.1.2.2 & Aquatic vertebrate & 16 & 366 & 694 & 316 & 45.5\% & 60.5\% & 70.2\%  \\
~ &  1.1.2.3 & Bird & 59 & 1,937 & 2,568 & 2,140 & \bf 83.3\% & 64.3\% & 74.4\%  \\
~ &  1.1.2.4 & Reptilian & 36 & 547 & 749 & 520 & \bf 69.4\% & 63.4\% & 73.2\%  \\
~ &  1.1.3 & Invertebrate & 61 & 1,317 & 1,792 & 1,284 & \bf 71.7\% & 61.6\% & 73.0\%  \\
\cmidrule[0.25pt]{1-10}
\parbox[t]{2mm}{\multirow{9}{*}{\rotatebox[origin=c]{90}{Vit-Base}}}  &  1 & Organism & 410 & 9,390 & 13,952 & 11,835 & \bf 84.8\% & 45.6\% & 55.6\%  \\
~ &  1.1.2.1.1 & Primate & 20 & 475 & 824 & 498 & \bf 60.4\% & 37.3\% & 49.3\%  \\
~ &  1.1.2.1.2 & Hoofed mammal & 17 & 419 & 490 & 224 & 45.7\% & 42.0\% & 50.6\%  \\
~ &  1.1.2.1.3 & Feline & 13 & 319 & 409 & 209 & \bf 51.1\% & 50.6\% & 61.9\%  \\
~ &  1.1.2.1.4 & Canine & 130 & 2,502 & 5,308 & 4,550 & \bf 85.7\% & 52.5\% & 64.0\%  \\
~ &  1.1.2.2 & Aquatic vertebrate & 16 & 366 & 477 & 234 & 49.1\% & 49.5\% & 61.8\%  \\
~ &  1.1.2.3 & Bird & 59 & 1,937 & 1,821 & 1,093 & \bf 60.0\% & 31.6\% & 41.1\%  \\
~ &  1.1.2.4 & Reptilian & 36 & 547 & 754 & 475 & \bf 63.0\% & 51.9\% & 59.2\%  \\
~ &  1.1.3 & Invertebrate & 61 & 1,317 & 1,685 & 1,023 & \bf 60.7\% & 48.5\% & 57.6\%  \\
\cmidrule[0.25pt]{1-10}
\parbox[t]{2mm}{\multirow{9}{*}{\rotatebox[origin=c]{90}{Vit-Large}}}  &  1 & Organism & 410 & 9,390 & 9,018 & 7,736 & \bf 85.8\% & 52.4\% & 62.0\%  \\
~ &  1.1.2.1.1 & Primate & 20 & 475 & 493 & 370 & \bf 75.1\% & 55.2\% & 63.5\%  \\
~ &  1.1.2.1.2 & Hoofed mammal & 17 & 419 & 324 & 155 & 47.8\% & 53.4\% & 59.9\%  \\
~ &  1.1.2.1.3 & Feline & 13 & 319 & 302 & 162 & \bf 53.6\% & 53.0\% & 61.3\%  \\
~ &  1.1.2.1.4 & Canine & 130 & 2,502 & 3,609 & 3,192 & \bf  88.4\% & 56.2\% & 68.3\%  \\
~ &  1.1.2.2 & Aquatic vertebrate & 16 & 366 & 316 & 155 & 49.1\% & 63.9\% & 71.5\%  \\
~ &  1.1.2.3 & Bird & 59 & 1,937 & 977 & 621 & \bf 63.6\% & 40.9\% & 49.4\%  \\
~ &  1.1.2.4 & Reptilian & 36 & 547 & 428 & 285 & \bf 66.6\% & 52.1\% & 61.7\%  \\
~ &  1.1.3 & Invertebrate & 61 & 1,317 & 1,154 & 760 & \bf 65.9\% & 59.2\% & 65.3\%  \\
\cmidrule[1pt]{1-10}
\end{tabular}
\label{tbl:organism_per_model}
\end{table}

\clearpage
\begin{table}[htbp!] 
\centering
\caption{For the adversarial examples that achieved model-to-model transferability and that have been created with \textbf{PGD} and \textbf{CW}, intra-collection misclassifications and misclassifications into the top-\{3,5\} prediction classes in the target models are provided for each model employed in this study ($1$st column). The results for the adversarial examples are grouped into collections according to the classes of their source image origins. The results are provided for a number of collections that lie under the \textbf{Artifact} sub-tree.}
\scriptsize
\begin{tabular}{lllccccccc}
\cmidrule[0.25pt]{1-10}
\parbox[t]{2mm}{\multirow{4}{*}{\rotatebox[origin=c]{90}{Model}}} & 
\multirow{4}{*}{\shortstack{Hierarchy}} & 
\multirow{4}{*}{\shortstack{Collection}} & 
\multirow{4}{*}{\shortstack{Classes\\in collection}} & 
\multirow{4}{*}{\shortstack{Source\\images\\in collection}} & 
\multirow{4}{*}{\shortstack{Adversarial\\examples\\originating\\from collection}} & 
\multicolumn{2}{c}{\multirow{3}{*}{\shortstack{Intra-collection\\misclassifications}}} &
\multicolumn{2}{c}{\multirow{3}{*}{\shortstack{Misclassification\\into top-K\\classes}}} \\
~ & ~ & ~ \\
~ & ~ & ~ \\
\cmidrule[0.25pt]{7-10}
~ & ~ & ~ & ~ & ~ &  ~ & Count & \%  & Top-3 & Top-5 \\
\cmidrule[0.25pt]{1-10}
\parbox[t]{2mm}{\multirow{9}{*}{\rotatebox[origin=c]{90}{AlexNet}}}  &  2 & Artifact & 522 & 8,397 & 18,149 & 16,341 & \bf 90.0\% & 72.5\% & 83.8\%  \\
~ &  2.1.1.1 & Clothing & 49 & 670 & 1,790 & 833 & 46.5\% & 67.0\% & 80.2\%  \\
~ &  2.2 & Covering & 90 & 1,287 & 2,960 & 1,386 & 46.8\% & 68.4\% & 81.2\%  \\
~ &  2.3.1 & Container & 99 & 1,528 & 3,396 & 1,806 & \bf 53.2\% & 79.3\% & 86.8\%  \\
~ &  2.3.1.2 & Wheeled vehicle & 43 & 879 & 1,554 & 927 & \bf 59.7\% & 84.6\% & 92.9\%  \\
~ &  2.3.3 & Device & 125 & 1,901 & 4,099 & 1,385 & 33.8\% & 71.8\% & 83.8\%  \\
~ &  2.3.3.4 & Musical instrument & 26 & 427 & 915 & 402 & 43.9\% & 74.4\% & 86.0\%  \\
~ &  2.3.4 & Equipment & 37 & 738 & 1,778 & 355 & 20.0\% & 63.7\% & 79.6\%  \\
~ &  2.4 & Structure & 57 & 1,035 & 1,733 & 876 & \bf 50.5\% & 84.2\% & 91.6\%  \\
\cmidrule[0.25pt]{1-10}
\parbox[t]{2mm}{\multirow{9}{*}{\rotatebox[origin=c]{90}{SqueezeNet}}}  &  2 & Artifact & 522 & 8,397 & 35,748 & 32,165 & \bf 90.0\% & 58.7\% & 71.0\%  \\
~ &  2.1.1.1 & Clothing & 49 & 670 & 3,474 & 1,038 & 29.9\% & 58.8\% & 72.4\%  \\
~ &  2.2 & Covering & 90 & 1,287 & 5,963 & 2,240 & 37.6\% & 60.8\% & 73.6\%  \\
~ &  2.3.1 & Container & 99 & 1,528 & 6,041 & 3,061 & \bf 50.7\% & 60.4\% & 72.9\%  \\
~ &  2.3.1.2 & Wheeled vehicle & 43 & 879 & 2,958 & 1,646 & \bf 55.6\% & 66.5\% & 78.0\%  \\
~ &  2.3.3 & Device & 125 & 1,901 & 7,781 & 2,282 & 29.3\% & 57.7\% & 70.5\%  \\
~ &  2.3.3.4 & Musical instrument & 26 & 427 & 1,674 & 428 & 25.6\% & 62.5\% & 75.0\%  \\
~ &  2.3.4 & Equipment & 37 & 738 & 3,732 & 588 & 15.8\% & 47.4\% & 60.9\%  \\
~ &  2.4 & Structure & 57 & 1,035 & 3,344 & 1,573 & 47.0\% & 65.5\% & 76.6\%  \\
\cmidrule[0.25pt]{1-10}
\parbox[t]{2mm}{\multirow{9}{*}{\rotatebox[origin=c]{90}{VGG-16}}}  &  2 & Artifact & 522 & 8,397 & 20,329 & 18,204 & \bf 89.5\% & 52.9\% & 66.0\%  \\
~ &  2.1.1.1 & Clothing & 49 & 670 & 2,197 & 929 & 42.3\% & 50.2\% & 64.7\%  \\
~ &  2.2 & Covering & 90 & 1,287 & 3,729 & 1,822 & 48.9\% & 53.5\% & 67.4\%  \\
~ &  2.3.1 & Container & 99 & 1,528 & 3,272 & 1,758 & \bf 53.7\% & 55.8\% & 69.8\%  \\
~ &  2.3.1.2 & Wheeled vehicle & 43 & 879 & 1,221 & 784 & \bf 64.2\% & 69.5\% & 81.7\%  \\
~ &  2.3.3 & Device & 125 & 1,901 & 4,082 & 1,334 & 32.7\% & 51.4\% & 62.5\%  \\
~ &  2.3.3.4 & Musical instrument & 26 & 427 & 697 & 265 & 38.0\% & 55.4\% & 65.6\%  \\
~ &  2.3.4 & Equipment & 37 & 738 & 2,132 & 451 & 21.2\% & 47.0\% & 60.0\%  \\
~ &  2.4 & Structure & 57 & 1,035 & 1,833 & 759 & 41.4\% & 56.9\% & 68.4\%  \\
\cmidrule[0.25pt]{1-10}
\parbox[t]{2mm}{\multirow{9}{*}{\rotatebox[origin=c]{90}{DenseNet-121}}}  &  2 & Artifact & 522 & 8,397 & 14,699 & 12,978 & \bf 88.3\% & 60.5\% & 71.5\%  \\
~ &  2.1.1.1 & Clothing & 49 & 670 & 1,487 & 593 & 39.9\% & 56.6\% & 70.5\%  \\
~ &  2.2 & Covering & 90 & 1,287 & 2,699 & 1,239 & 45.9\% & 61.1\% & 73.1\%  \\
~ &  2.3.1 & Container & 99 & 1,528 & 2,566 & 1,317 & \bf 51.3\% & 69.9\% & 76.9\%  \\
~ &  2.3.1.2 & Wheeled vehicle & 43 & 879 & 1,122 & 678 & \bf 60.4\% & 76.9\% & 82.1\%  \\
~ &  2.3.3 & Device & 125 & 1,901 & 2,963 & 1,163 & 39.3\% & 61.5\% & 72.3\%  \\
~ &  2.3.3.4 & Musical instrument & 26 & 427 & 577 & 310 & \bf 53.7\% & 69.0\% & 78.5\%  \\
~ &  2.3.4 & Equipment & 37 & 738 & 1,246 & 346 & 27.8\% & 50.0\% & 63.0\%  \\
~ &  2.4 & Structure & 57 & 1,035 & 1,700 & 639 & 37.6\% & 63.5\% & 72.4\%  \\
\cmidrule[0.25pt]{1-10}
\parbox[t]{2mm}{\multirow{9}{*}{\rotatebox[origin=c]{90}{ResNet-50}}}  &  2 & Artifact & 522 & 8,397 & 12,887 & 11,576 & \bf 89.8\% & 57.7\% & 69.2\%  \\
~ &  2.1.1.1 & Clothing & 49 & 670 & 1,352 & 528 & 39.1\% & 63.1\% & 75.3\%  \\
~ &  2.2 & Covering & 90 & 1,287 & 2,376 & 1,112 & 46.8\% & 64.7\% & 76.4\%  \\
~ &  2.3.1 & Container & 99 & 1,528 & 2,210 & 1,156 &\bf  52.3\% & 62.3\% & 73.1\%  \\
~ &  2.3.1.2 & Wheeled vehicle & 43 & 879 & 911 & 589 & \bf 64.7\% & 73.4\% & 82.8\%  \\
~ &  2.3.3 & Device & 125 & 1,901 & 2,285 & 832 & 36.4\% & 55.7\% & 65.9\%  \\
~ &  2.3.3.4 & Musical instrument & 26 & 427 & 341 & 180 & \bf 52.8\% & 64.5\% & 73.6\%  \\
~ &  2.3.4 & Equipment & 37 & 738 & 1,181 & 242 & 20.5\% & 48.9\% & 62.1\%  \\
~ &  2.4 & Structure & 57 & 1,035 & 1,501 & 561 & 37.4\% & 57.5\% & 68.9\%  \\
\cmidrule[0.25pt]{1-10}
\parbox[t]{2mm}{\multirow{9}{*}{\rotatebox[origin=c]{90}{Vit-Base}}}  &  2 & Artifact & 522 & 8,397 & 10,771 & 9,359 & \bf 86.9\% & 47.1\% & 56.2\%  \\
~ &  2.1.1.1 & Clothing & 49 & 670 & 1,042 & 454 & 43.6\% & 48.9\% & 60.5\%  \\
~ &  2.2 & Covering & 90 & 1,287 & 1,918 & 837 & 43.6\% & 48.1\% & 59.3\%  \\
~ &  2.3.1 & Container & 99 & 1,528 & 1,893 & 923 & 48.8\% & 49.3\% & 58.4\%  \\
~ &  2.3.1.2 & Wheeled vehicle & 43 & 879 & 906 & 490 & \bf 54.1\% & 54.3\% & 62.4\%  \\
~ &  2.3.3 & Device & 125 & 1,901 & 1,998 & 738 & 36.9\% & 40.8\% & 48.7\%  \\
~ &  2.3.3.4 & Musical instrument & 26 & 427 & 341 & 135 & 39.6\% & 46.3\% & 53.4\%  \\
~ &  2.3.4 & Equipment & 37 & 738 & 892 & 247 & 27.7\% & 44.1\% & 54.8\%  \\
~ &  2.4 & Structure & 57 & 1,035 & 1,488 & 539 & 36.2\% & 45.9\% & 53.6\%  \\
\cmidrule[0.25pt]{1-10}
\parbox[t]{2mm}{\multirow{9}{*}{\rotatebox[origin=c]{90}{Vit-Large}}}  &  2 & Artifact & 522 & 8,397 & 7,374 & 6,458 & \bf 87.6\% & 54.3\% & 63.2\%  \\
~ &  2.1.1.1 & Clothing & 49 & 670 & 668 & 285 & 42.7\% & 53.0\% & 64.7\%  \\
~ &  2.2 & Covering & 90 & 1,287 & 1,283 & 546 & 42.6\% & 52.5\% & 63.0\%  \\
~ &  2.3.1 & Container & 99 & 1,528 & 1,401 & 680 & 48.5\% & 56.7\% & 66.6\%  \\
~ &  2.3.1.2 & Wheeled vehicle & 43 & 879 & 616 & 331 & \bf  53.7\% & 63.1\% & 71.1\%  \\
~ &  2.3.3 & Device & 125 & 1,901 & 1,228 & 501 & 40.8\% & 49.5\% & 57.3\%  \\
~ &  2.3.3.4 & Musical instrument & 26 & 427 & 211 & 115 & \bf 54.5\% & 59.7\% & 66.8\%  \\
~ &  2.3.4 & Equipment & 37 & 738 & 509 & 150 & 29.5\% & 52.1\% & 62.7\%  \\
~ &  2.4 & Structure & 57 & 1,035 & 1,200 & 402 & 33.5\% & 55.1\% & 63.6\%  \\

\cmidrule[1pt]{1-10}
\end{tabular}
\label{tbl:artifact_per_model}
\end{table}

\end{document}